%% file: main.tex
\documentclass[twoside]{article}

\usepackage[accepted]{aistats2026}
\usepackage[utf8]{inputenc} 
\usepackage[T1]{fontenc}    
\usepackage{hyperref}       
\usepackage{url}            
\usepackage{booktabs}       
\usepackage{amsfonts}       
\usepackage{nicefrac}       
\usepackage[expansion=true, protrusion=false]{microtype}      
\usepackage{xcolor}         

\usepackage{natbib}
\usepackage{amssymb, amsmath,amsthm}
\usepackage{algorithm}
\usepackage{algorithmic}
\usepackage{paralist}
\usepackage{multirow}

\usepackage{enumerate}
\usepackage{makeidx}  
\usepackage{mathtools}
\usepackage{xspace}
\usepackage{mathrsfs}
\usepackage{graphicx} 
\usepackage{epstopdf}
\usepackage{bm}
\usepackage{bbm}
\usepackage{upgreek}
\usepackage{cleveref}
\usepackage{cancel}
\usepackage{empheq}
\usepackage{pifont}
\usepackage{subcaption}
\usepackage{stfloats} 

\theoremstyle{plain}
\newtheorem{theorem}{Theorem}[section]

\newtheorem{lemma}[theorem]{Lemma}

\theoremstyle{definition}
\newtheorem{definition}[theorem]{Definition}

\theoremstyle{remark}

\newcommand{\cmt}[1]{}

\newcommand{\ours}{{CADSR}\xspace}

\input{./emacscomm.tex}

\begin{document}

\twocolumn[
\aistatstitle{Complexity-Aware Deep Symbolic Regression with Robust Risk-Seeking Policy Gradients}

\aistatsauthor{ Zachary Bastiani $^{1,2}$ \And Robert M. Kirby $^{1,2}$ \And  Jacob Hochhalter$^{3}$ \And Shandian Zhe $^{1}$ }
\aistatsaddress{$^{1}$ Kahlert School of Computing \quad
    $^{2}$ Scientific Computing and Imaging Institute \\
    $^{3}$ Department of Mechanical Engineering, University of Utah
}




\runningauthor{Zachary Bastiani, Robert M. Kirby, Jacob Hochhalter, Shandian Zhe}
]

\input{abstract}
\input{intro}
\input{method}

\input{related}

\input{expr}

\input{conclusion}

\newpage
\input{acknowledgments}

\emergencystretch=1em
\bibliographystyle{apalike}
\bibliography{references}

\section*{Checklist}

\begin{enumerate}

  \item For all models and algorithms presented, check if you include:
  \begin{enumerate}
    \item A clear description of the mathematical setting, assumptions, algorithm, and/or model. [Yes]
    \item An analysis of the properties and complexity (time, space, sample size) of any algorithm. [Yes]
    \item (Optional) Anonymized source code, with specification of all dependencies, including external libraries. [Yes]
  \end{enumerate}

  \item For any theoretical claim, check if you include:
  \begin{enumerate}
    \item Statements of the full set of assumptions of all theoretical results. [Yes]
    \item Complete proofs of all theoretical results. [Yes]
    \item Clear explanations of any assumptions. [Yes]     
  \end{enumerate}

  \item For all figures and tables that present empirical results, check if you include:
  \begin{enumerate}
    \item The code, data, and instructions needed to reproduce the main experimental results (either in the supplemental material or as a URL). [Yes]
    \item All the training details (e.g., data splits, hyperparameters, how they were chosen). [Yes]
    \item A clear definition of the specific measure or statistics and error bars (e.g., with respect to the random seed after running experiments multiple times). [Yes]
    \item A description of the computing infrastructure used. (e.g., type of GPUs, internal cluster, or cloud provider). [Yes]
  \end{enumerate}

  \item If you are using existing assets (e.g., code, data, models) or curating/releasing new assets, check if you include:
  \begin{enumerate}
    \item Citations of the creator If your work uses existing assets. [Yes]
    \item The license information of the assets, if applicable. [Not Applicable]
    \item New assets either in the supplemental material or as a URL, if applicable. [Yes]
    \item Information about consent from data providers/curators. [Not Applicable]
    \item Discussion of sensible content if applicable, e.g., personally identifiable information or offensive content. [Not Applicable]
  \end{enumerate}

  \item If you used crowdsourcing or conducted research with human subjects, check if you include:
  \begin{enumerate}
    \item The full text of instructions given to participants and screenshots. [Not Applicable]
    \item Descriptions of potential participant risks, with links to Institutional Review Board (IRB) approvals if applicable. [Not Applicable]
    \item The estimated hourly wage paid to participants and the total amount spent on participant compensation. [Not Applicable]
  \end{enumerate}

\end{enumerate}

\clearpage
\onecolumn
\appendix
\input{appendix}



\end{document}

%% file: emacscomm.tex
\newcommand{\h}{{\bf h}}

\newcommand{\x}{{\bf x}}

\newcommand{\A}{{\bf A}}

\newcommand{\C}{{\bf C}}

\renewcommand{\H}{{\bf H}}

\newcommand{\K}{{\bf K}}

\newcommand{\Q}{{\bf Q}}

\newcommand{\Scal}{{\mathcal{S}}}

\newcommand{\V}{{\bf V}}

\newcommand{\wh}[1]{{\widehat{#1}}}

\newcommand{\whatR}{\widehat{R}}

\newcommand{\0}{{\bf 0}}

\newcommand{\ben}{\begin{enumerate}}
\newcommand{\een}{\end{enumerate}}

\newcommand{\ie}{{\textit{i.e.,}}\xspace}

\newcommand{\eg}{{{\textit{e.g.},}}\xspace}

%% file: abstract.tex
\begin{abstract}
    We propose a novel deep symbolic regression (DSR) approach to enhance the robustness and interpretability of data-driven mathematical expression discovery.
    Existing DSR methods are built on recurrent neural networks, solely guided by data fitness, and potentially meet tail barriers that can zero out the policy gradient, causing inefficient model updates.
    To address these issues, we design a decoder-only architecture that performs attention in the frequency domain and introduce a dual-indexed position encoding to conduct layer-wise generation. 
    Second, we propose a Bayesian information criterion (BIC)-based reward function that can automatically adjust the trade-off between expression complexity and data fitness, without the need for explicit manual tuning.
    Third, we develop a  ranking-based weighted policy update method that eliminates the tail barriers and enhances training effectiveness.
    Extensive benchmarks and systematic experiments demonstrate the  advantages of our approach.
    We have released our implementation at \url{https://github.com/ZakBastiani/CADSR}.
\end{abstract}

%% file: intro.tex
\section{Introduction}
Symbolic regression (SR)~\citep{schmidt_distilling_2009, jobin_global_2019, rudin_stop_2019} is an important research direction for achieving interpretability in machine learning.  Given a dataset that records the input and output of a complex system of interest, symbolic regression seeks to discover a simple, concise equation that reveals the system's underlying mechanism --- thereby enhancing our understanding and ensuring the model's reliability.

Genetic programming (GP)~\citep{koza_genetic_1994, randall_bingo_2022, burlacu_operon_2020} has long been the dominant approach for symbolic regression. However, GP is known to be computationally expensive and time-consuming due to its evolutionary nature. 
Deep Symbolic Regression (DSR)~\citep{petersen_deep_2019} and its variants~\citep{tenachi2023deep, jiang2024vertical} represent a recent breakthrough by training a recurrent neural network (RNN) to generate expressions  from data efficiently. While DSR has demonstrated success across many SR benchmarks, the RNN-based architecture can struggle with capturing long-range dependencies and is prone to vanishing gradients~\citep{hochreiter_vanishing_1998}, especially in large expression trees. In addition, DSR methods typically rely on data-fitting rewards, which can lead to overly complex expressions and overfitting --- particularly in noisy settings. Furthermore, due to the usage of the reward difference as the weights in the policy gradients, DSR takes the risk of meeting tail barriers, \ie zero policy gradients, which can result in inefficient model updates. 


To address these issues, we introduce a robust, complexity-aware deep symbolic regression method (\ours). It is worth noting that \ours and the DSR framework aim to learn a data-specific expression generator from scratch, rather than utilize a pretrained supervised model mapping datasets directly to expressions~\citep{valipour_symbolicgpt_2021,shojaee_transformer-based_2023}. Our major contributions are as follows. 
 
    \noindent \textbf{Expression Model:} We design a decoder-only architecture that performs attention in the frequency space. We apply the discrete cosine transform to the embeddings to obtain their frequency representations. High-frequency components --- often attributed to noise, outliers, or excessive parameter perturbations --- are removed, and attention is computed over the resulting low-frequency components. An inverse transform is then applied to recover updated embeddings in the original space. 
     Next, to enhance the model's ability to capture positional information within the expression tree, we introduce a dual-indexed positional encoding based on both the depth and horizontal locations of each token. We use a breadth-first search (BFS) strategy to generate the expression tree layer by layer, which offers both computational efficiency and implementation simplicity.
    
    \noindent\textbf{Reward Design:} We propose a Bayesian Information Criterion (BIC)-based reward function that computes the model evidence to evaluate the trade-off between the expression complexity and data fit.  This enables the learning process to explicitly optimize the trade-off between interpretability and accuracy,  avoiding overly complex expressions that tend to overfit --- particularly in the presence of noise. BIC is grounded in Bayesian model selection theory~\citep{wasserman2000bayesian} and is closely related to the principle of Minimum Description Length (MDL)~\citep{Rissanen1978}, making it a principled and robust criterion. Moreover, BIC introduces no additional hyperparameters, allowing for automatic trade-off adjustment to maximize model evidence, and eliminating the need for manual tuning.
    

    \noindent \textbf{Policy Gradient:} We propose a novel risk-seeking policy gradient method, integrated with Group Relative Policy Optimization (GRPO)~\citep{shao2024deepmath}. Unlike DSR, which uses reward differences as gradient weights, our approach employs a ranking-based linear weighting scheme to perform step-wise reward mapping. This design not only preserves distinctions among the top-ranked candidates but also  avoids any tail barriers or partial tail barriers in the gradient updates. As a result, our method can effectively utilize top-performing samples for model updates, minimizing inefficient or overly exploratory behavior. Furthermore, integration with GRPO enables repeated and reliable optimization steps for each batch of sampled expressions within a trust region, thereby enhancing training effectiveness while reducing sample complexity.
    
    \noindent \textbf{Experiments:} We evaluated  \ours  on the standard SR benchmark, a practical fracture mechanical application, and many ablation studies. In addition to DSR, we compared with seventeen other popular and/or state-of-the-art SR methods and several commonly used machine learning approaches. The performance of \ours in both symbolic discovery and prediction accuracy is consistently among the best. In particular, the symbolic discovery rate of \ours is the highest when data includes significant noises. 
    In all the cases, \ours generates the most interpretable expressions, while maintaining a high level of accuracy.  
    \ours  outperforms the most comparable model, DSR, in all categories showing that it is a direct improvement. Extensive ablation studies further demonstrate the effectiveness of each component of our method.

%% file: method.tex
\section{Background}
Given a set of input and output examples collected from the target system, denoted as $\mathcal{D}=\{(\x_i, y_i)\}_{i=1}^N$, symbolic regression aims to identify a concise expression that characterizes the input-output relationship, such as $y = \sin(2\pi x_1) + \cos(2\pi x_2)$.
Deep symbolic regression (DSR)~\citep{petersen_deep_2019} discovers equations via an RNN-based reinforcement learning approach~\citep{sutton2018reinforcement}, which can be broken down into four parts: environment, actor, reward, and policy.

The environment is designed to be the creation of an expression tree that represents a specific equation.
Expression trees are directed trees where each node holds a token from the available list of operations and variables, \eg \{$+$, $-$, $\times$, $x_1$\}.
Expression trees are built by selecting nodes in a preorder traversal of the trees. 
These trees are many-to-one mappings to the mathematical expressions, which can increase the search space but prevent generating invalid expressions.

The actor is an RNN that predicts a categorical distribution of the available tokens for each node in the expression tree based on the  hidden state of the RNN and the sibling and parent of the current node. 
Each token is randomly sampled from the categorical distribution. 
Additional rules are applied to the 
sampling process
to prevent the selection of redundant operations or variables.

The reward function and policy 
drive the actor to explore and exploit  the complex environment. In DSR, the reward function is a direct measurement of the data fitness of the generated expression,
\begin{align}
    R(\tau) = \frac{1}{1 + \text{NRMSE}}, \label{eq:dsr-reward}
\end{align}
where $\tau$ denotes the expression, NRMSE represents the normalized root-mean-square error, and is defined as $\text{NRMSE} = \frac{1}{\sigma_y}\sqrt{\frac{1}{n}\sum_{i=1}^{n}(y_i - \tau(\x_i))^2}$ where $\sigma_y$ is the standard deviation of the training output in the dataset. 
DSR applies a risk-seeking policy to update the actor model according to its best predictions. Specifically, at each step, DSR samples a batch of expressions, ranks their rewards, and selects the top $\alpha\%$ expressions to compute a policy gradient, 
\begin{align}
	\begin{split}
		\nabla_\theta J_{\text{risk}}(\theta ; \alpha) = \frac{1}{\alpha B/100} \sum\nolimits_{i=1}^{B} [R(\tau^{(i)}) - R_{\alpha}] \\
		\cdot \textbf{1}_{R(\tau^{(i)}) \ge R_{\alpha}}\nabla_{\theta} \log(p(\tau^{(i)} | \theta)), \label{risk_seeking_gradient}
	\end{split}
\end{align}
where $B$ is the batch size, $\mathbf{1}_{(\cdot)}$ is an indicator function,  
$\tau^{(i)}$ is the $i$-th expression in the batch, $\theta$ denotes the RNN parameters, $\log(p(\tau^{(i)} | \theta))$ is the probability of $\tau^{(i)}$ being sampled by the current RNN, 
$R_\alpha$ is the $1-\alpha/100$ quantile of the rewards in the batch. 
Accordingly, all the equations below top $\alpha\%$ will not influence the update of the actor.
This policy enables the actor to generate lower-performing equations without negatively impacting its overall performance, as we only care about the top performed expressions. This  allows for more unrestrained exploration and targeted exploitation of the top performers. 


\section{Method}

\subsection{Decoder-Only Expression Generator}
We first design a decoder-only transformer architecture for expression generation. Unlike RNNs, which rely on a single hidden state to summarize information across all previous nodes, making them susceptible to vanishing gradients and poor at capturing long-range dependencies~\citep{hochreiter_vanishing_1998},  transformers compute dependencies explicitly between all nodes through the attention mechanism. 
This allows them to prevent gradient vanishing and effectively capture various short- and long-range dependencies.
\begin{figure}
	\centering
	\includegraphics[width=0.3\textwidth]{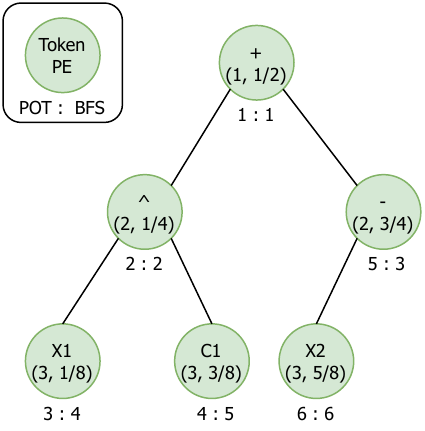}
	\caption{\small Expression tree for $y = x_{1}^{c_1} + \sin(x_{2})$. POT and BFS denote the node order for preorder traversal and breadth-first-search, respectively.}
	\label{fig:expression_tree_example}
\end{figure}

\textbf{Dual-Indexed Position Encoding.} 
We use one-hot encoding to represent each token, such as operations, variables, and constants, and adopt an autoregressive approach to generate each node in the expression tree. Instead of assigning a single pre-order index to represent each node, we introduce a dual-indexed position encoding (DPE) to more effectively capture the tree's structural information. Specifically,  given a particular node, we consider both its depth $d$, and horizontal  location $h$, in the tree. To align the horizontal locations of the nodes across different layers, we propose the following design. Denote the horizontal location of the parent node by $h_p$,  if the node is the left child, we assign its horizontal position as $h = h_p - h_p/2^d$, otherwise we assign $h= h_p + h_p/2^d$. In this way, the parent node will be in between its two children horizontally, which naturally reflects the tree structure. The horizontal position for the root node is set to $1/2$. See Fig.~\ref{fig:expression_tree_example} for an illustration.  We develop a recursive algorithm to efficiently calculate the horizontal positions of all the nodes, as listed in Appendix Algorithm~\ref{alg:PE_initization}. Given $d$ and $h$, we  construct a 2D-dimensional encoding, 
\begin{align}
	\text{DPE}(d,h)_{2i} &= \sin(\frac{d}{10000^{(4i/D)}}), \\
	\text{DPE}(d,h)_{2i+1} &= \cos(\frac{d}{10000^{(4i/D)}}), \\
	\text{DPE}(d,h)_{D+2j} &= \sin(\frac{h}{10^{(4j/D)}}), \\
	\text{DPE}(d,h)_{D+2j+1} &= \cos(\frac{h}{10^{(4j/D)}}) \label{eq:positional_encoding} 
\end{align}
where $0 \le i, j \le \lfloor D/2 \rfloor$. The embedding of each tree node is constructed as the positional encoding plus the one-hot encoding of the token. 

\textbf{Robust Attention in Frequency Space.} To capture a diverse range  of complex, short- and long-range dependencies among tree nodes\cmt{  for new node generation}, we construct multiple self-attention layers. Inspired by the recent DCT-Former~\citep{Scribano_2023} developed for large language models, we propose performing attention in the frequency domain to improve both learning robustness and efficiency. Let  the embeddings of the current  tree nodes be denoted as $\H = [\h_0, \ldots, \h_{N-1}]^\top$, where each $\h_n \in \mathbb{R}^r$ ($0 \le n \le N-1$). We treat these embeddings as samples of a function with $r$-dimensional outputs at $N$ discrete input locations. We then apply a Discrete Cosine-Transform (DCT) to extract their frequency-domain representations $\wh{\H} = [\wh{\h}_0, \ldots, \wh{\h}_{N-1}]^\top$, where 
\begin{align}
\wh{\H} = \C \H, \;\;\; C_{k,n} = \alpha_k \cos\left[\frac{\pi}{N}\left(n+\frac{1}{2}\right)k\right]. \label{eq:dct}
\end{align}
Here, $0 \le k, n \le N-1$, and the normalization constant $\alpha_k$ is defined as $\alpha_k = \sqrt{1/N}$ for $k=0$, and $\alpha_k = \sqrt{2/N}$ otherwise. 
The DCT decomposes the function into a weighted sum of cosine basis functions oscillating at different frequencies $k=0, \ldots, N-1$. Each $\wh{\h}_k \in \mathbb{R}^r$ represents the coefficient of the cosine basis at frequency $k$.  One advantage of using the DCT is that the resulting frequency-domain representation remains entirely in the real-valued domain. In contrast, applying a Fourier transform would yield complex-valued representations, requiring special handling to avoid numerical issues. To improve robustness and efficiency during training, we apply a low-pass filter by removing the high-frequency components, which are often associated with noise, outliers in data, or excessive perturbations to the model parameters. This filtering helps to denoise the representations and stabilize learning. Specifically, we retain only the first $M$ frequency components, denoted as $\wh{\H}_{1:M} = [\wh{\h}_0, \ldots, \wh{\h}_{M-1}]^\top$, where $M$ is a tunable hyperparamter. We perform attention on $\wh{\H}_{1: M}$:
\begin{align}
\wh{\A} 
   &= \text{Attention}(\wh{\H}_{1:M}; \Q, \K, \V) \notag \\
   &= \text{softmax}\Big(\tfrac{(\wh{\H}_{1:M} \Q)(\wh{\H}_{1:M} \K)^\top}{\sqrt{r}}\Big)
      (\wh{\H}_{1:M} \V). \label{eq:dct-atten}
\end{align}
We perform this self-attention in the frequency domain multiple times to fully capture the rich, complex dependencies of the frequency components. Then we transform the representation back to the original domain using inverse DCT:
\begin{align}
	\H \leftarrow \C^\top [\wh{\A};\0], 
\end{align} 
where the zero-padding restores the original dimensionality by appending $N-M$  frequency components.
To predict the token distribution for the next node, we apply a linear layer followed by a softmax function. The overall architecture of our model is illustrated in Appendix Fig. \ref{fig:cadsr_arch}.

\textbf{Expression Sampling.} Starting from the root node, we generate the expression tree in a layer-wise, breadth-first-search (BFS) order, auto-regressively expanding the nodes layer by layer. This generation scheme aligns naturally with our dual-indexed positional encoding, enabling effective integration of the spatial relationships between the previously generated nodes and the current target node for token sampling. Moreover, since the tree grows in a fixed layer-by-layer fashion, the ordering of previously generated nodes remains unchanged throughout the process. As a result, we only need to perform vectorization once prior to sampling, making the approach both computationally efficient and straightforward to implement.

Given the predicted token distribution, we first mask out invalid tokens and then sample from the remaining candidates. Once the sampled tokens form a valid expression, the generation process terminates and returns the resulting expression. The complete sampling procedure is summarized in Algorithm~\ref{alg:expr-tree-sampling} in the Appendix.

\subsection{BIC Reward Function}

Interpretability is a central motivation for symbolic regression. However, when using only data fit as the reward signal --- such as in DSR --- the model tends to produce overly complex expressions that overfit the data, especially in the presence of noise, which is common in real-world applications. This overfitting undermines the interpretability of the learned expressions. To address this issue, we adopt the Bayesian Information Criterion (BIC)~\citep{schwarz1978estimating} to define a new reward function. BIC is theoretically grounded in Bayesian model selection \citep{wasserman2000bayesian} and closely related to the Minimum Description Length (MDL) principle~\citep{Rissanen1978}, effectively serving as its approximation.

As a principled and robust criterion, BIC computes the model evidence by integrating out the model parameters, which in essence evaluates the trade-off between the model complexity and data fit in terms of generalizability. In our design, we define the model complexity $k$ as the sum of the number of nodes in the expression tree and the number of constant tokens. This inclusion serves two purposes: (1) Since the values of constant tokens are unknown a priori and must be estimated from data, they effectively act as model parameters, introducing additional degrees of freedom; (2) penalizing the number of constant tokens discourages their excessive use, which would otherwise increase the computational cost of optimizing their values. The BIC reward is defined as: 
\begin{align} 
	\text{BIC}(\tau) &= k \log(S) - 2 \log(p(\tau)), \label{eq:bic} 
\end{align} 
where $\log(p(\tau)) = \sum_{i=1}^S \log \mathcal{N}(y_i|\tau(\x_i), \sigma^2)$ is the Gaussian log-likelihood, $\sigma^2$ is the variance of the training outputs, and $S$ is the number of training samples. To accommodate more complex data distributions, $p(\tau)$ can be replaced with a Student-$t$ likelihood or a mixture of Gaussians. Notably, the BIC reward introduces no additional tuning hyperparameters. This allows for automatic balancing between model complexity and data fit, enabling the actor to directly optimize for the expression's generalizability without manual intervention.

\subsection{Robust Risk-Seeking Policy}
While our BIC reward~\eqref{eq:bic} accounts for expression complexity, its value is unbounded and can grow arbitrarily large. As a result, directly applying it in a risk-seeking policy framework, as done in DSR, can introduce unpredictable high variance and lead to highly unstable learning. Even after normalization, a few outlier expressions can dominate the reward distribution, causing the majority of well-performing samples to contribute little or nothing to the policy update.

To mitigate this issue, a commonly used strategy, which is also adopted in DSR, is to introduce a continuous mapping that maps the reward value to a bounded domain. For example, DSR uses the mapping $f(z) = \frac{1}{1 + z}$ where $z \in [0, \infty]$, which maps the unbounded NRMSE (in $[0, \infty]$) to the range $(0, 1]$; see \eqref{eq:dsr-reward}. Another choice can be the sigmoid function, $s(z)=\frac{1}{1 + e^{-z}}$ that maps from $(-\infty, \infty)$ to $(0, 1)$. In this way, we can control the variance and improve learning stability.
However, this strategy introduces another challenge. That is, the reinforcement learning  can encounter a \textit{tail barrier}, defined as follows.
\begin{definition}[Tail barrier]
     Let \begin{math}\alpha \in [0, 1)\end{math}. A risk seeking policy meets a \begin{math}\alpha\end{math}-tail barrier if the top $\alpha$\% rewards of the sampled actions (\eg expression trees) are all equal to $R_\alpha$.
\end{definition}  
\begin{lemma}\label{lem:tail}
    Given any continuous mapping $f$ that can map unbounded reward function values to a bounded domain (\eg $(-\infty, \infty) \rightarrow [0, 1]$), suppose the reward function is continuous, there always exists a set of distinct rewards values that numerically create a tail barrier in the risk-seeking policy.
\end{lemma}
The proof is given in Appendix Section~\ref{sect:proof-tail}. In practice, since we often use continuous reward functions (\eg NRMSE or Gaussian likelihood) and reward mappings, there is a risk of encountering the tail barrier.  
From~\eqref{risk_seeking_gradient}, we can see that the tail barrier can incur \textit{zero} policy gradient, since all the top $\alpha$\% rewards are identical to $R_\alpha$, leading to a zero weight for every gradient. 
As a consequence, the actor model would not have any effective updates according to the feedback from the selected expressions (top performers). In DSR, the RNN model will be updated only from an entropy bonus term~\citep{petersen_deep_2019}, and henceforth the learning starts to explore wildly.

To address this challenge, we introduce a ranking-based linear weighting scheme to construct a step-wise reward mapping, 
\begin{align}
    f(z) = \lambda \cdot \text{ReLU}\left( 1- \tfrac{|\{\tau^{(i)}: R(\tau^{(i)}) > R(z) \}|}{ \alpha B/100}\right) \label{eq:step-map}
\end{align}
where $z$ denotes an arbitrary  expression  sampled by the actor,  $|\cdot|$ represents the size of the set, and $\lambda>0$ is a scaling constant. This formulation ensures that expressions ranked within the top $\alpha\%$ receive positive mapped rewards, with higher-ranked expressions assigned larger values of $f(z)$, while all other expressions receive a reward of zero. Consequently, only top-performing expressions contribute to the gradient update, and lower-ranked ones are effectively ignored.
The ranked-based weighting of our policy produces smoother updates, prevents pathological reward spikes, and improves training stability by removing tail barriers.
Our risk-seeking policy gradient is therefore: 
\begin{align}
    \nabla J_{\text{risk}}(\theta ; \alpha) \approx \frac{100}{\alpha B} \sum_{i=1}^{B} f(\tau^{(i)}) \nabla_{\theta} \log p(\tau^{(i)} | \theta). \label{eq:new-risk-seeking}
\end{align}

\begin{lemma}\label{lem:robust}
By using the step function~\eqref{eq:step-map} for reward mapping, the policy gradient with our BIC reward, as shown in \eqref{eq:new-risk-seeking}, is unbiased and will not encounter any tail barrier. 
\end{lemma}
The proof is given in Appendix Section~\ref{sect:proof-robust}.

\noindent \textbf{Group Relative Policy Optimization}.
Inspired by the recent success in large language model reasoning, we adapt our policy to conceptually align with the Group Relative Policy Optimization (GRPO) framework~\citep{shao2024deepmath} to further enhance the learning stability and efficiency. Equation \ref{eq:new-risk-seeking} provides a group-relative advantage function which is integrated into a PPO- and TRPO-based policy ~\citep{schulman2017proximalpolicyoptimizationalgorithms}. Specifically, among the top $\alpha\%$ candidates, we only use the tokens from those expressions within a trust-region determined by the relative variation in trajectory probability as compared to the earlier model version: 
\begin{align*}
\nabla \wh{J}(\theta ; \alpha) 
   =& \tfrac{100}{\alpha B} 
      {\textstyle \sum_{i=1}^{B} \sum_{j=1}^{|\tau^{(i)}|}}
      f(\tau^{i})\cdot\\ 
      &\nabla_\theta\min\!\Big[g(\tau^i, j),
      \operatorname{clip}\big(g(\tau^i, j), 1\pm\epsilon\big)\Big], \\
g(\tau^i, j) 
   =& \tfrac{p(\tau_{j}^{(i)}|\tau_{<j}^{(i)}, \theta)}
           {p(\tau_{j}^{(i)}|\tau_{<j}^{(i)}, \theta_{\text{old}})},
\end{align*}
where $\theta_{\text{old}}$ denotes the model's parameters at the beginning of the current epoch, $\epsilon \in (0, 1)$ controls the size of the trust-region, and $|\tau^{(i)}|$ is the number of tokens in the expression.  
The trust region can prevent violent model updates caused by outlier samples, as the clipping operator sets $g(\tau^i, j) > 1 + \epsilon$ to a constant, thereby yielding a gradient of zero.
To further improve the learning stability, we introduce a penalty term that penalizes model changes that are too aggressive. Our final policy update is:  
\begin{align}
	\begin{split}
		\nabla J_{\text{GRPO}}(\theta; \alpha) = \nabla \wh{J}(\theta ; \alpha) - \frac{100}{\alpha B}  \sum\nolimits_{i=1}^{B} \sum\nolimits_{j=1}^{|\tau^{(i)}|}  \\
		\quad  \beta\cdot  \nabla \text{KL}[p(\tau_{j}^{(i)}| \tau_{<j}^{(i)}, \theta)\| p(\tau_{j}^{(i)}| \tau_{<j}^{(i)}, \theta_{\text{ref}})], \label{eq:grpo-update}
	\end{split}
\end{align} 
where $\text{KL}(\cdot\|\cdot)$ is the Kullback-Leibler divergence,  $\theta_{\text{ref}}$ denotes the parameters of the reference model, which is chosen from a previous epoch, and $\beta>0$ controls  the penalty strength.  We use an experience replay buffer~\citep{mnih2013playingatarideepreinforcement} to merge the historically best performing $\alpha B/100$ expressions for each epoch. Our approach is summarized in Appendix Algorithm~\ref{alg:ours}.

%% file: related.tex
\section{Related Work}
Since its invention, genetic programming (GP) has served as a dominant approach for symbolic regression~\citep{koza_genetic_1994}. Recent representative works include Bingo~\citep{randall_bingo_2022},  Operon~\citep{burlacu_operon_2020},  GP-GOMEA~\citep{virgolin_improving_2021}, and others. \citet{bomarito_automated_2023} introduced a variant GP that leverages factional Bayes factor to indicate the most likely expression for noisy datasets. 

The DSR framework~\citep{petersen_deep_2019} uses reinforcement learning to train an RNN-based expression generator from a given measurement dataset. Recent extensions by~\citet{tenachi2023deep, jiang2024vertical} have incorporated domain knowledge into DSR by enforcing physics-unit constraints or generating rules within a vertical discovery space for vector symbolic regression. The DSR framework and our method can be regarded as \textit{unsupervised} learning: only numerical measurements are provided, with no ground-truth expressions available. The goal, in fact, is to discover or generate those underlying expressions. 
DySymNet ~\citep{li2024neuralguideddynamicsymbolicnetwork} is another reinforcement learning method that trains an RNN controller to generate the structure of an EQL network ~\citep{Kim_2021}. The generated EQL-network is fit to the training data with sparsity regularization applied to the network's weights. DySymNet's reward signal leverages the network's fit quality and complexity to select for compact, high-quality network structures. DySymNet hybridizes DSR-like search methods with EQL networks to improve accuracy and reduce complexity relative to EQL. DySymNet differs from \ours through its application of reinforcement learning to improve EQL-network structure.

An alternative line of work aims to train a foundation model that can map numerical datasets directly to symbolic expressions~\citep{biggio2021neural, kamienny_end--end_2022, valipour_symbolicgpt_2021, vastl_symformer_2022,li2023transformerbased}. These approaches adopt a \textit{supervised} learning paradigm, where the training data consist of input-output pairs: numerical measurements as inputs and their corresponding ground-truth symbolic expressions as outputs. The objective is to learn the underlying mapping between the two.
These approaches typically rely on an encoder-decoder transformer architecture, where the encoder layers extract information from the numerical measurements, and the decoder layers integrate this information to generate expressions. 
 While promising, such foundation models are expensive to train and highly sensitive to data preparation, which often involves fabricating large-scale synthetic datasets. More importantly, due to the absence of a data-specific search mechanism, these methods may struggle with generalization --- particularly when applied to out-of-distribution data~\citep{kamienny2023deepgenerativesymbolicregression}.
To address this, recent research has explored coupling pretrained foundation models with data-specific search or planning mechanisms to improve expression discovery. Notable examples include TPSR~\citep{shojaee_transformer-based_2023}, DGSR~\citep{holtdeep23}, and DGSR-MCTS~\citep{kamienny2023deepgenerativesymbolicregression}. TPSR integrates a pretrained model within a Monte Carlo Tree Search (MCTS)~\citep{browne2012survey} framework, using the model to guide token selection and search tree expansion via an upper confidence bound (UCB) heuristic. The MCTS reward function includes a complexity penalty, controlled by a tunable hyperparameter. 

Other notable efforts include uDSR~\citep{landajuela2022a}, an ensemble framework combining multiple symbolic regression methods such as GP and DSR, and model-free approaches~\citep{sun2023symbolic, xu2024reinforcement} that rely solely on MCTS or hybrid strategies combining MCTS and GP for symbolic expression search.

%% file: expr.tex
\section{Experiments} \label{sec:numerical_exp}
For evaluation, we examined \ours on the comprehensive and well-known SRBench dataset~\citep{la_cava_contemporary_2021}, and then conducted an ablation study to assess the effectiveness of each individual component. Finally, we showcased the effectiveness of \ours through an application to fracture mechanics.
\subsection{Overall Performance}
\begin{figure*}[!h]
    \centering
    \includegraphics[width=0.95\textwidth]{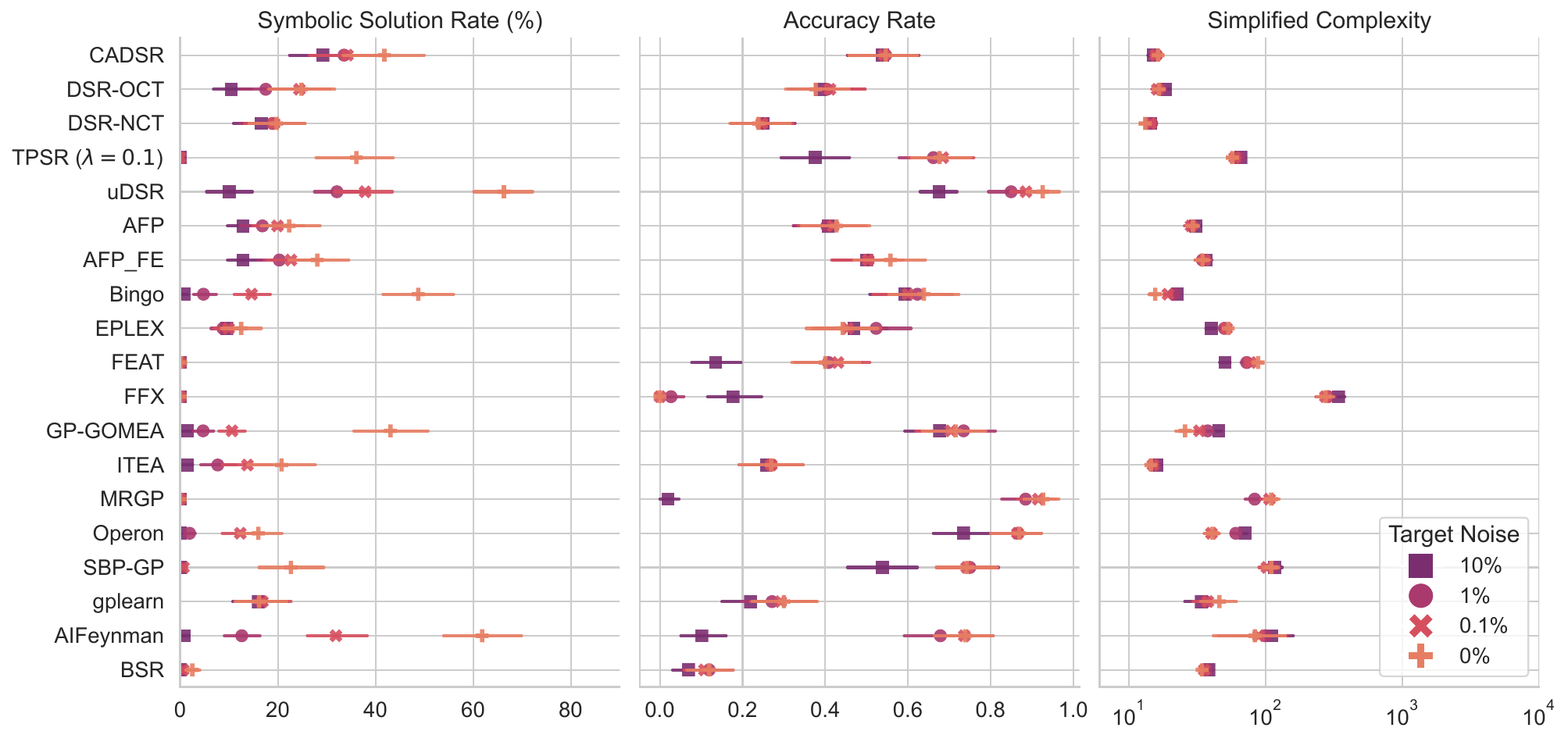}
    \caption{\small Symbolic regression performance on 133 SRBench problems with known solutions. Error bars denote a $95\%$ confidence region. The numerical values are reported in Appendix Table~\ref{tab:results_symbolic_regression}. }
    \label{fig:ground_truth_sr_bench}
\end{figure*}

In SRBench, we first tested on the 133 problems with known solutions. 
We ran eight trials for each problem at the four \textit{noise} levels: 0\%, 0.1\%, 1\%, and 10\%. We ran \ours in a large computer cluster, for which we set a time limit of 4 hours for each trial. We deployed the trials on NVIDIA A40 GPUs. 
The maximum number of epochs is set to 600 without early termination.  
 In each epoch, we sampled a batch of 1000 expressions to compute the policy gradient. We used ADAM optimization with an initial learning rate of 1E-4. The full list of hyperparameters of our method is provided in Appendix Table~\ref{tab:hyperparameters}. We compared with 17 popular and/or state-of-the-art SR methods in terms of Symbolic Solution Rate (\%), Accuracy Rate, and Simplified Complexity, which are standard metrics for SR evaluation. These SR baselines include DSR~\citep{petersen_deep_2019}, Bingo~\citep{randall_bingo_2022},  GP-GOMEA~\citep{virgolin_improving_2021}, ITEA~\citep{10.1162/evco_a_00285}, TPSR~\citep{shao2024deepmath}, BSR~\citep{jin2020bayesian}, AIFeynman~\citep{udrescu_ai_2020}, AFP\_FE \citep{schmidt_distilling_2009},  among others.   
The majority of these are genetic programming methods with a few notable exceptions: DSR is deep reinforcement learning, TPSR leverages a pretrained foundation model~\citep{kamienny_end--end_2022} to conduct Monte-Carlo tree search over expressions, 
BSR is an MCMC method with a prior placed on the tree structure, and AIFeynmen is a divide-and-conquer method that breaks the problem apart by hyper-planes and fits with polynomials.
The results of the competing methods are retrieved from the public SRBench report~\citep{la_cava_contemporary_2021} and from published resources. For TPSR, we used the recommended setting of $\lambda = 0.1$, as suggested by the authors, which provides the best trade-off between expression accuracy and complexity.  The DSR results reported in the SRBench report correspond to the variant with No Constant Tokens (DSR-NCT), which may not fully capture the method's performance. To address this, we additionally evaluated an enhanced version of DSR by introducing constant tokens into the token library and optimizing them before each policy update, referred to as DSR-OCT (DSR with Optimized Constant Tokens). The results of all methods are presented in Fig.~\ref{fig:ground_truth_sr_bench}.

Overall, \ours shows strong performance in symbolic discovery as measured by Symbolic Solute Rate. In particular, when data includes significant noise (10\%), \ours achieves the best solution rate, showing that our method is more \textit{robust} to noise than all the competing methods. Meanwhile, the simplified complexity of our discovered expressions is among the lowest. This together shows that  our method, with the BIC reward design, not only can find simpler and hence more interpretable expressions, but also is more resistant to data noise. 
The Accuracy Rate shows the ratio for which the method is able to discover an equation with $R^2 > 0.999$. \ours has a competitive accuracy rate with AFP\_FE, and Bingo, while outperforming DSR with and without constant optimization.
The slightly better methods, such as TPSR and GP-GOMEA, however, generate lengthier and more complex expressions, which lack interpretability and are much far away from the ground truth expression. It is worth noting that \ours outperforms DSR in both Symbolic Solution Rate (\%) and Accuracy Rate, showing an improvement on both expression discovery and prediction accuracy. 
When data is noise-free, AIFeynman, uDSR, and Bingo shows better Symbolic Solution Rate than \ours. This might be because: AIFeynman tends to use polynomials to construct the expressions, which match most of the ground-truth; Bingo as a genetic programming approach, uses evolution operators to sample new expressions, which might explore more broadly; uDSR is an ensemble approach using AIFeynman, genetic programming, and DSR and thus can achieve higher symbolic accuracy. However, in the presence of noise, the performance of these methods deteriorates largely and immediately falls behind \ours, demonstrating \ours 's superior robustness to noisy data

\begin{figure}[!hb]
	\centering
	\includegraphics[width=0.4\textwidth]{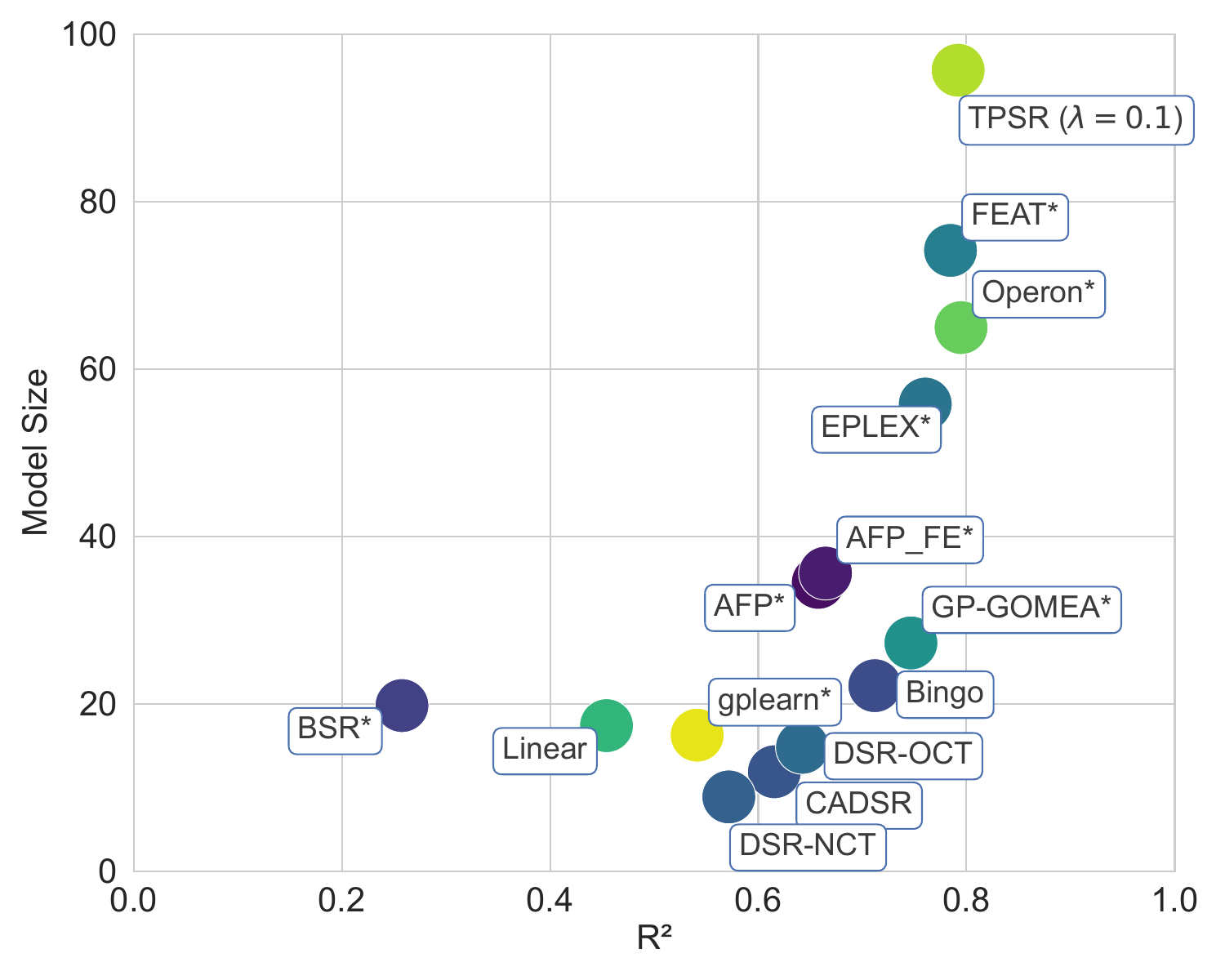}
	\caption{\small Pareto front of each method in 120 black-box problems of SRBench; the true solutions are unknown. Numerical metrics are reported in Appendix Table~\ref{tab:blackbox_problems_table}.}\label{fig:blackbox-srbench}
\end{figure} 
Next, we tested with the 120 black-box problems in SRBench. These problems consists of real-world
scientific and engineering problems spanning domains such as health informatics, technology, environmental science, and economics. 
Since the true solutions are \textit{unknown}, we examined the Pareto front of all the methods in Model Size  \textit{vs.} $R^2$ Test Rank. 
As shown in Fig.~\ref{fig:blackbox-srbench}, \ours lies on the Pareto frontier
, meaning \ours is among the best in terms of the trade-off between the model size  (expression complexity) and $R^2$ score (prediction accuracy). It is interesting to see that \ours lies between DSR-NCT and DSR-OCT. On the contrary, methods like Operon and TPSR, typically generates way more complex expressions yet with smaller prediction error.
It shows that our method can push the best trade-off toward more interpretability due to BIC penalizing complexity, which can be important in practice.

\noindent\textbf{Runtime.} In Appendix Figure~\ref{fig:srbench_time_comparison}, we provide a runtime comparison across all the methods. \ours achieves an average runtime of approximately 4 hours over all problems, which is faster than {DCT-OCT} and slower than {DCT-NCT}. Note that DCT-NCT does not involve any constant optimization. On average, Bingo requires around 6 hours, while AIFeynman takes approximately 8 hours. Overall, these results confirm the training efficiency of \ours.

\subsection{Ablation Study}
To evaluate the effectiveness of each component of \ours, we ran ablations on the Feynman dataset from SRBench, which is composed of 119 problems. The numerical values of the metrics are given in Appendix \ref{appendix:ablation_results}.

\begin{figure}[t]
	\centering
	\includegraphics[width=0.8\linewidth]{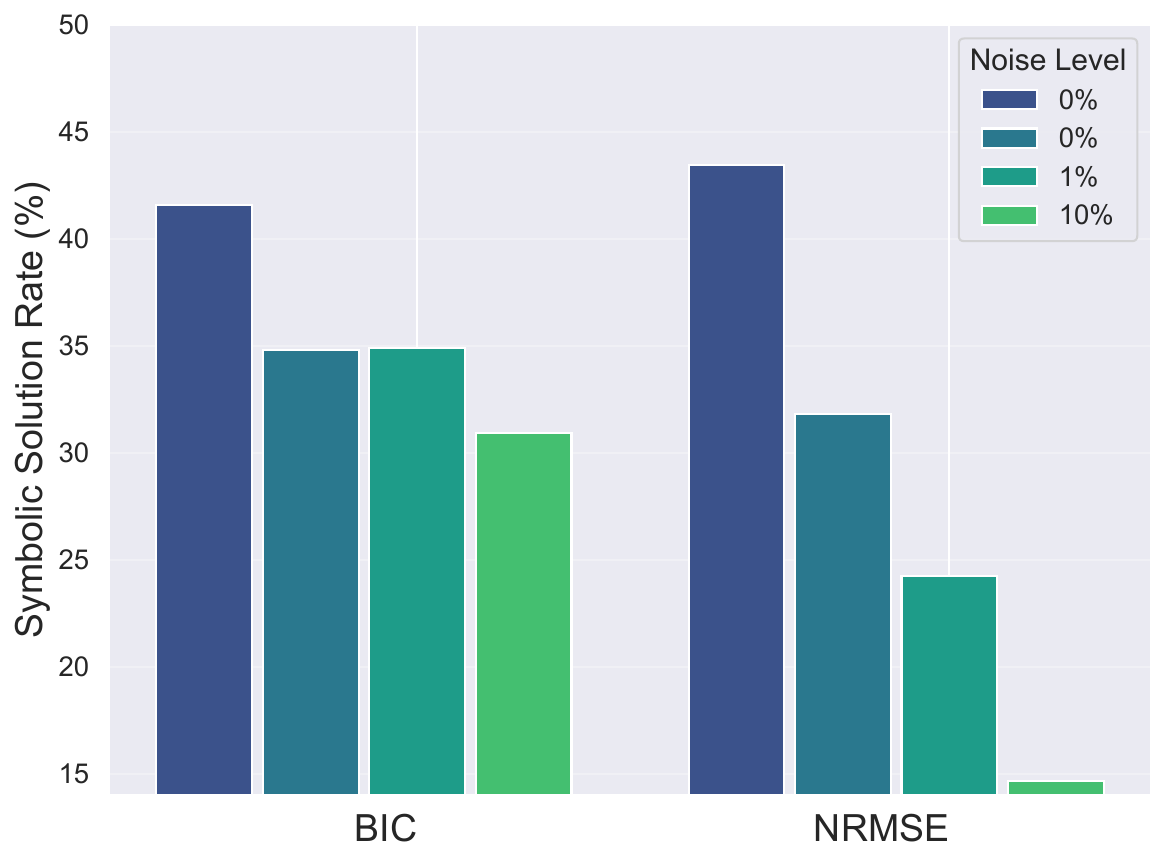}
	\caption{\ours performance the standard NRMSE reward function.}
	\label{fig:bic_nmse_ablation}
\end{figure}

\textbf{BIC Reward Function}: To validate our BIC-based reward function, we first compared it against the standard NRMSE-based reward (see~\eqref{eq:dsr-reward}) across varying noise levels. As shown in Figure~\ref{fig:bic_nmse_ablation}, while the BIC reward performs slightly worse under zero noise, it substantially outperforms the NRMSE-based reward at all other noise levels. Notably, at a 10\% noise level, the BIC reward improves upon the NRMSE-based reward by {111\%} in Symbolic Solution Rate.

\begin{figure*}[t]
	\centering
	\begin{tabular}{cc}
		\begin{subfigure}{0.5\linewidth}
			\includegraphics[width=\linewidth]{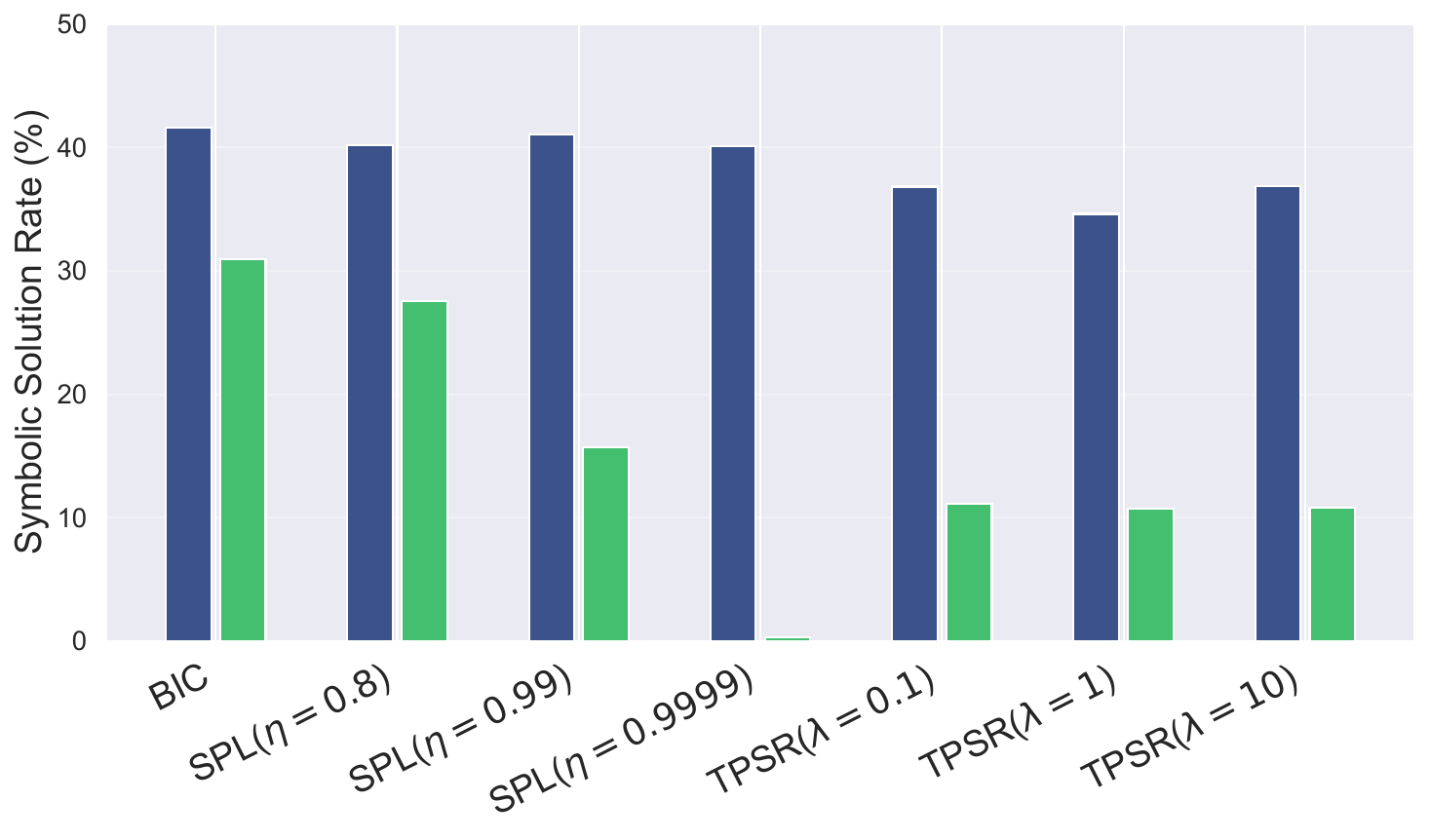}
			\caption{\small BIC \textit{vs.} SPL \textit{vs.} TPSR reward functions}
			\label{fig:bic_regularizers_ablation}
		\end{subfigure} &
		\begin{subfigure}{0.395\linewidth}
			\includegraphics[width=\linewidth]{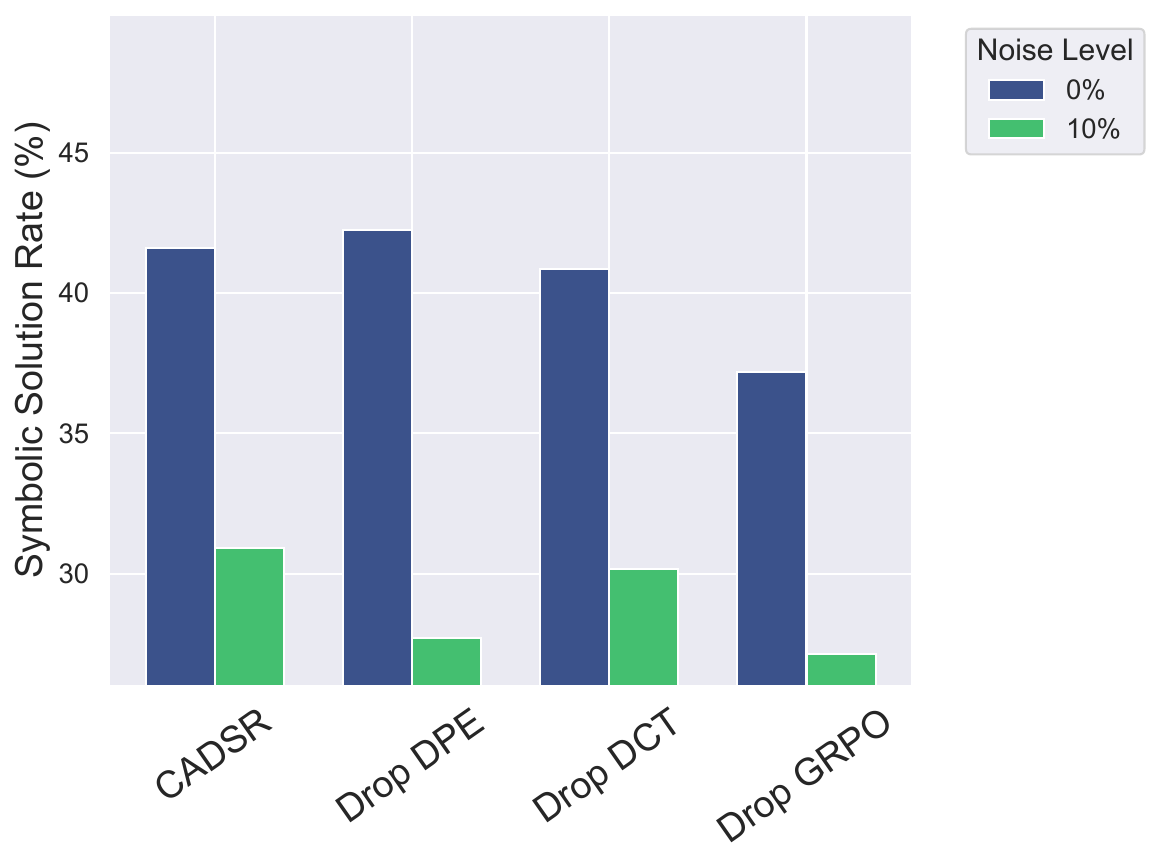}
			\caption{\small Individual component ablations.}
			\label{fig:dct_ablation}
		\end{subfigure}
	\end{tabular}
	\caption{\small Symbolic Solution Rate of \ours with (a) alternative reward function and (b) individual component removes on Feynman dataset.}
\end{figure*}

In addition, we evaluated our BIC reward against two recently proposed alternatives: the SPL~\citep{sun2023symbolic} and TPSR~\citep{shojaee_transformer-based_2023} reward functions, which both incorporate an explicit trade-off hyperparameter to balance expression complexity and data fit --- denoted by $\eta$ in SPL and $\lambda$ in TPSR. Definitions of these rewards are provided in~\eqref{eq:spl} and~\eqref{eq:tpsr} in the Appendix. We performed the comparison at noise levels of 0\% and 10\%, testing a range of values for the trade-off hyperparameters. As shown in Figure~\ref{fig:bic_regularizers_ablation}, our BIC reward --- which requires no manually tuned trade-off parameter --- consistently outperforms both alternatives across all settings. Specifically, the SPL reward achieves its highest Symbolic Solution Rate at $\eta = 0.99$ (for 0\% noise) and $\eta = 0.8$ (for 10\% noise), yet still falls short by 0.5\% and 3\%, respectively, compared to our BIC reward. The TPSR reward reaches its peak at $\lambda = 0.1$ for both noise levels, trailing our method by {5\% and 20\%}. These results highlight the strength of our BIC-based approach, particularly in noisy settings, and confirm its effectiveness in balancing complexity and data fidelity without requiring parameter tuning
 
\textbf{Other Components}:
Next, we conducted an ablation study by individually removing other components from \ours to evaluate their contributions to overall performance. Specifically, we replaced the DCT attention layer (see~\eqref{eq:dct} and~\eqref{eq:dct-atten}) with a standard attention mechanism, substituted the dual-indexed position encoding (DPE) with a 1D positional encoding, and replaced GRPO with a standard risk-seeking policy gradient. The comparative results are presented in Figure~\ref{fig:dct_ablation}.
As shown, removing each component generally led to a performance drop. The only exception was replacing DPE with 1D positional encoding, which resulted in a slight improvement of 0.5\% in the noise-free setting; however, at the 10\% noise level, performance declined by 3\%. The DCT attention module not only improved the Symbolic Solution Rate compared to standard attention, but also reduced expression sampling time (see \ref{appendix:dct_runtimes} for runtime details). GRPO had a substantial impact: its removal decreased the symbolic discovery rate by 4.5\%, and also slowed convergence --- with the final expression discovered at an average of epoch 257, compared to epoch 168 with GRPO. Detailed convergence plots are provided in Appendix~\ref{appendix:grpo}. 

\subsection{Fracture Mechanics}
Finally, we applied \ours to a critical problem in fracture mechanics --- generating accurate and interpretable expressions for predicting crack initiation in microstructures ~\citep{Spear_2018}. 
The dataset consists of six grain-level features used to predict the Fatigue Indicator Parameters (FIPs) ~\citep{Hansen_2024}, which are known to be correlated with crack initiation. FIPs lack a closed-form solution and typically require computationally expensive simulations, so we aim to identify surrogate expressions that can quickly approximate FIPs ~\citep{McDowell_2010}. We compare \ours against DSR, GP-GOMEA, TPSR, and DySymNet, running eight trials per method with access to both CPU and GPU resources. The details about the dataset and the running of each method is provided in Appendix Section~\ref{sect:fracture}. 

The $R^2$ score and complexity of discovered expressions are reported in Table~\ref{tab:fracture_mechanics_results}. In Appendix Table~\ref{tab:expr-fracture}, we list the top expressions found by each method. As we can see, CADSR achieves the best trade-off between accuracy and interpretability. It consistently produces concise, interpretable expressions with strong predictive performance. In contrast, TPSR fails to generate accurate expressions, likely due to the numerical domain of the problem being outside its foundational model's training set.
DSR achieves performance comparable to CADSR but yields slightly longer and less accurate expressions. GP-GOMEA provides marginal improvements in accuracy but at the cost of significantly higher complexity, producing expressions that were on average 50.75 tokens longer than \ours and effectively uninterpretable. 
DySymNet produced significantly complex expression without improving the $R^2$ score. We suspect that DySymNet's performance is affected by its 8-hour runtime limit, as converging to its optimal expression may require additional computation.
Overall, CADSR prioritizes simplicity and interpretability while maintaining accuracy, enabling engineers to better understand how cracks initiate.

\begin{table}
\centering
\small 
\begin{tabular}{lrr}
\hline
Method   & $R^2$ & Complexity \\
\hline
CADSR     & 0.626 & 22.25 \\
DSR       & 0.616 & 24.63 \\
TPSR      & 0.046 & 40.75 \\
GP-GOMEA  & 0.636 & 73.00 \\
DySymNet  & 0.456 & 1697.88 \\
\hline
\end{tabular}
\caption{\small Symbolic regression performance for predicting crack initiation.}
\label{tab:fracture_mechanics_results}
\end{table}

%% file: conclusion.tex
\section{Conclusion}

We have presented \ours, a new symbolic regression approach based on reinforcement learning. On standard SR benchmark problems, \ours shows promising performance. The ablation study confirms the effectiveness of each component of our method. The Fracture Mechanics problem enforces \ours application for real world problems.  Nonetheless, our current work has two limitations. First, the implementation is inefficient, especially for expression optimization, as each expression optimization is time intensive, causing a slow training process. A potential solution is to leverage a hybrid optimization approach, using a fast gradient-based optimizer (e.g., Adam, BFGS) initially, followed by refinement with LM. Thus, a hybrid approach could yield the same accuracy and stable constant optimization of LM while reducing total runtime. Second, we lack an early stopping mechanism to reduce the training cost further and prevent useless exploration. In the future, we plan to address these limitations.


%% file: acknowledgments.tex
\section{Acknowledgements}

ZB and JH acknowledge supports from Air Force Research Lab STTR Phase under project number 250504306.  SZ acknolwedges support from  NSF CAREER Award IIS-2046295,  NSF OAC-2311685 (Elements: A Convergent Physics-based and Data-driven Computing Platform for Building Modeling), NSF DMS-2529112 (Collaborative Research: MATH-DT: Computationally efficient hypercomplex variable-based sensitivity methods for rapid Digital Twin model updating)

This work used Delta at the National Center for Supercomputing Applications through allocation CIS250317 from the Advanced Cyberinfrastructure Coordination Ecosystem: Services \& Support (ACCESS) program \cite{boerner2023access}, which is supported by U.S. National Science Foundation grants \#2138259, \#2138286, \#2138307, \#2137603, and \#2138296.

%% file: appendix.tex
\section*{Appendix}

\section{Algorithms}
Below, we show the algorithms for expression sampling and positional encoding generation. Note that for expression sampling, we over-sample expressions so that we can return a high number of unique ones. If not enough unique expressions exist, then we begin to allow duplicates to fill out our batch size requirements.
\begin{algorithm}[!ht]
\small 
    \caption{\small Expression Tree Sampling}\label{alg:expr-tree-sampling}
    \textbf{input} Number of expressions to sample $B$; oversampling scalar $\gamma>1$, maximum tree-node number $\nu$\\
    \textbf{output} A set of expressions $\mathcal{T}$
    \begin{algorithmic}[1]
    \STATE $\mathcal{T} \gets \text{ExpressionTrees}(\gamma B)$ 
    \COMMENT{Creates $\gamma B$ empty expression trees}
    \WHILE{$i < \nu$}
        \STATE $V_{\mathcal{T}} \gets \text{Inputs}(\mathcal{T})$
        \COMMENT{Fetching the input embeddings of all the expression trees}
        \STATE $S \gets p(V_{\mathcal{T}} | \theta)$ 
        \COMMENT{Predicting categorical distributions from the transformer}
        \STATE $S \gets R(S)$
        \COMMENT{Applying rules to each distribution}
        \STATE $K \gets P(\cdot | S)$
        \COMMENT{Sampling from the categorical distribution to obtain tokens}
        \STATE $\mathcal{T}_{i} \gets K$
        \COMMENT{Adding the new tokens into the expression trees}
    \ENDWHILE
    \STATE $\mathcal{T} = \text{Unique}(\mathcal{T}, B)$
    \COMMENT{Take the first $B$ Unique expression trees}
    \STATE \textbf{return} $\mathcal{T}$
    \end{algorithmic}
\end{algorithm}

\begin{algorithm}[!ht]
\small 
\caption{\small Dual Indexed Position Encoding (DPE) Generation}\label{alg:PE_initization}
    \textbf{input} An expression tree $\tau$
    \begin{algorithmic}[1]
        \STATE $\tau$.root\_node.depth = 1
        \STATE $\tau$.root\_node.horizontal = 1/2
        \STATE PositionEncodingInformation($\tau$.root\_node)
    \end{algorithmic}
\end{algorithm}

\begin{algorithm}[!ht]
\small
\caption{\small PositionEncodingInformation}\label{alg:PE_Recusive}
    \textbf{input} Current node
    \begin{algorithmic}[1]
    \IF{node has left}
        \STATE node.left.depth = node.depth + 1
        \STATE node.left.horizontal =  node.horizontal - $1/(2^{\text{node.left.depth}})$
        \STATE PositionEncodingInformation(node.left)
    \ENDIF
    \IF{node has right}
        \STATE node.right.depth =  node.depth + 1
        \STATE node.right.horizontal =  node.horizontal + $1/(2^{\text{node.right.depth}})$
        \STATE PositionEncodingInformation(node.right)
    \ENDIF
    \end{algorithmic}
\end{algorithm}

\begin{algorithm}[!t]
	\caption{Complexity-Aware Deep Symbolic Regression (\ours)}\label{alg:tdsr}
	\textbf{input} Learning rate $l$; risk factor $\alpha$; batch size $B$; coefficients $\lambda>0$; steps per epoch $C$; epochs per reference $G$; number of epochs $T$\\
	\textbf{output} The best equation $\tau^*$
	\begin{algorithmic}[1]
		\STATE Initialize transformer with parameters $\theta$ \\
		\WHILE{ $i < T$}
		
		\IF{$i \text{ mod}(G)=0$}
		\STATE $\theta_{\text{ref}} \gets \theta$ \quad\quad
		\COMMENT{Set the reference weights to the current weights}
		\ENDIF
		\STATE $\theta_{\text{old}} \gets \theta$ \quad\quad
		\COMMENT{Set the old weights to the current weights}
		\STATE $\mathcal{T} \gets \{\text{OptimizeConstants}(\tau^{(i)}) : \tau^{(i)} \sim p(\cdot | \theta)\}_{i=1}^{B}$
		\quad\quad\COMMENT{Sample $B$ expressions from the transformer actor and optimize the values of constant tokens}
		\STATE $\mathcal{R} \gets \{\text{BIC}(\tau^{(i)})\}_{i=1}^{B}$
		\quad\quad \COMMENT{Calculate the reward for each expression using~\eqref{eq:bic}}
		\STATE $\mathcal{R_{\alpha}} \gets (1-\alpha/100)\text{-quantile of }\mathcal{R}$
		\STATE $\mathcal{T} \gets \{\tau^{(i)} : \mathcal{R}(\tau^{(i)} \ge \mathcal{R_{\alpha}} \}$
		\quad\quad\COMMENT{Pick top $\alpha\%$ expressions}
		\STATE $\mathcal{T} \gets \mathcal{T} \cup \mathcal{T}_{\text{historical}}$
		\quad\quad\COMMENT{Merge with the historical top performing expressions}
		\FOR{$j=1$ to $C$}
		\STATE $\theta \gets \theta + l \cdot (\nabla_{\theta} \
		J_{\text{GRPO}}+ \lambda_{\mathcal{H}} \cdot \text{Entropy-Bonus})$
		\quad\quad\COMMENT{Compute the policy gradient using~\eqref{eq:grpo-update}; the entropy bonus term comes from the original DSR.}
		\STATE $\textbf{if} \text{ max } \mathcal{R} > \mathcal{R}(\tau^*) \textbf{ then } \tau^* \gets \tau^{(\text{argmax } \mathcal{R} \in \mathcal{T})}$
		\quad\quad\COMMENT{Update the best equation}
		\ENDFOR
		\STATE $\mathcal{T}_{\text{historical}} \gets \text{Top $\alpha\%$ of the expressions in $\mathcal{T}$}$
		\quad\quad\COMMENT{Update the experience replay buffer}
		\STATE $i \gets i+1$
		\ENDWHILE
		\STATE \textbf{return} $\tau^*$
	\end{algorithmic}\label{alg:ours}
\end{algorithm}

\section{Model Details}
\begin{figure}
    \centering
    \includegraphics[width=0.75\textwidth]{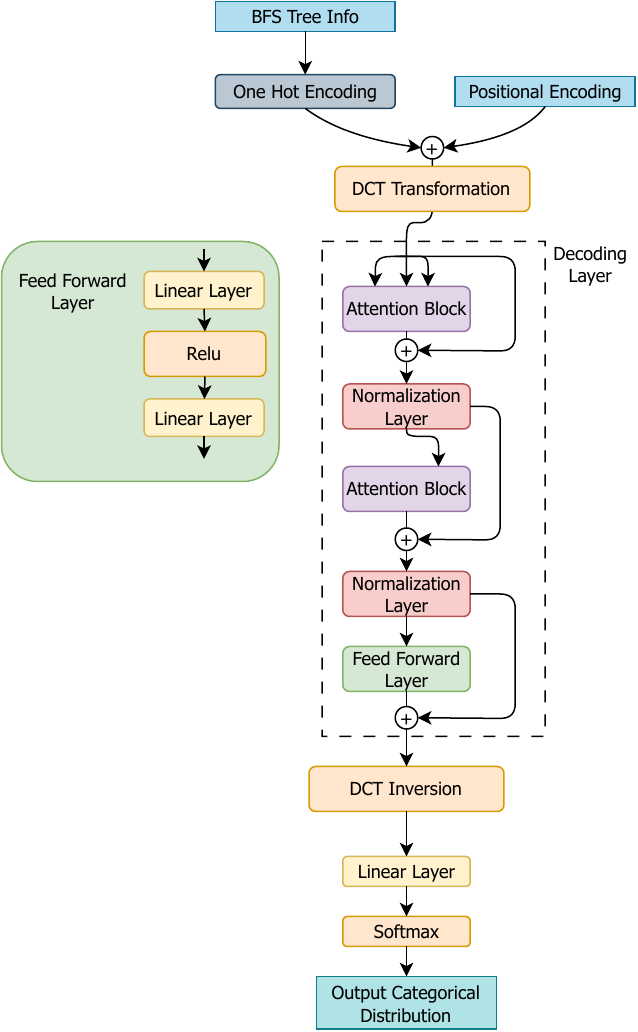}
    \caption{\small The architecture of the transformer actor in CADSR.}
    \label{fig:cadsr_arch}
\end{figure}

Table~\ref{tab:hyperparameters} and Fig.~\ref{fig:cadsr_arch} show the comprehensive hyperparameter settings and the architecture of the transformer used in \ours. Note that we use the Levenberg–Marquardt algorithm \citep{Levenberg_1944} to optimize the constant tokens for each discovered equation and used Adam \citep{kingma2017adammethodstochasticoptimization} to optimize the model.
\begin{table} [!ht]
    \centering
    \caption{\small Hyperparameter settings of \ours.}
    \small
    \begin{tabular}{c|c|c}
    \hline\hline
        \textit{Hyperparameter} & SRBench & Fracture Mechanics\\ \hline \hline
        Variables & \{1, $c$ (Constant Token), $x_{i}$\} &  \{1, $c$ (Constant Token), $x_{i}$\} \\ \hline
        Unary Functions & \{sin, cos, log, $\sqrt{(\cdot)}$, exp\} &\{sin, cos, log, $\sqrt{(\cdot)}$, exp, sqrt, tan, square\} \\ \hline
        Binary Functions & \{+, -, *, /, $\hat{}$ \} & \{+, -, *, /, $\hat{}$ \} \\ \hline
        Batch Size & 1000 & 2000 \\ \hline
        Risk Seeking Percent ($\alpha$) & $5\%$  & $5\%$ \\ \hline
        Learning Rate & 1E-4 & 1E-4 \\ \hline
        Max Depth & 32 & 32 \\ \hline
        Oversampling &  2 & 3  \\ \hline
        Number of Epochs & 600 & 500 \\ \hline
        Policy &  BIC &  BIC\\ \hline 
        $\lambda$ & 0.2 & 0.2\\ \hline
        Entropy Coefficient $\lambda_\mathcal{H}$ & 0.005 & 0.0005\\ \hline
        Encoder Number & 0 & 0\\ \hline
        Decoder Number & 1 & 2 \\ \hline
        Number of Heads & 1 & 3 \\ \hline
        Feed Forward Layers Size & 2048 & 2048 \\ \hline
        $\beta$ & 0.01 & 0.1 \\ \hline
        $\epsilon$ & 0.2 & 0.2 \\ \hline
        Embedding Dim & 10 & 15\\ \hline
        DCT Clip Dim & 8 & 12 \\ \hline
        $C$ & 5 & 5 \\ \hline
        $G$ & 5 & 5 \\ \hline
        CPU Count & 1 & 10
    \end{tabular}
    \label{tab:hyperparameters}
\end{table}

\begin{table}[!ht]
    \centering
    \small
    \caption{Hyperparameter settings for DSR}
    \begin{tabular}{c|c}
        \hline\hline
        \textit{Hyperparameter} & \textit{DSR} \\ \hline \hline
         Batch Size & 1000 \\ \hline
         Learning Rate & 0.0005 \\ \hline
         Entropy coefficient & 0.005 \\ \hline
         Risk Factor Percent &  $5\%$ \\ \hline
         RNN Type & LSTM \\ \hline
         Layer Number & 1  
    \end{tabular}

    \label{tab:my_label}
\end{table}

\newpage

\subsection{Sensitivity Analysis}

We conducted a compact sensitivity analysis of the key parameters introduced by \ours: DCT frequency cutoff $M$, trust region $\epsilon$, and KL Divergence regularizer $\beta$. We ran three trials for each setting on the Strogatz dataset with $0\%$ noise. Table \ref{tab:sensitivity_analysis} presents the results of this analysis with bolded columns referencing the hyperparameter setting used for the SRBench experiments.

For the DCT frequency cutoff, we found that removing high-frequency components is beneficial. Not clipping any frequencies led to a $21.4\%$ drop in solution rate. Interestingly, the model appears to depend strongly on the low-frequency domain: discarding the highest $60\%$ of frequencies ($M = 4$) produced only a modest performance change and increased the expression complexity by just 1.43 tokens compared with our default ($M = 8$), which suggests that DCT truncation acts primarily as a model regularizing mechanism.

The trust-region sweep indicates that the method relies on a sufficiently tight region to prevent unstable or erroneous updates. Increasing $\epsilon$ results in notable performance degradation across all metrics, while overly small $\epsilon$ values also reduce performance — likely because an excessively restrictive trust region slows or hinders policy improvement.

The KL penalty coefficient $\beta$ plays an influential role as Table \ref{tab:sensitivity_analysis} highlights the degradation in performance when altering $\beta$. Smaller values of $\beta$ decreased both accuracy and solution rate, presumably due to insufficient regularization and overly aggressive updates. Conversely, larger values of $\beta$ led to reductions in accuracy and symbolic recovery, consistent with over-regularization.

\begin{table}[t]
    \centering
    \caption{Sensitivity analysis for new parameters introduced by \ours}
    \label{tab:sensitivity_analysis}
    \small
    \begin{tabular}{lcccccccccccc}
        \toprule
        & \multicolumn{4}{c}{\textbf{DCT Cutoff ($\omega$)}} 
        & \multicolumn{4}{c}{\textbf{Trust Region ($\delta$)}}
        & \multicolumn{4}{c}{\textbf{KL Regularizer ($\lambda$)}} \\
        \cmidrule(lr){2-5} \cmidrule(lr){6-9} \cmidrule(lr){10-13}
        & 4 & 6 & \textbf{8} & 10 
        & 0.01 & \textbf{0.2} & 0.5 & 1.0 
        & 0.001 & 0.01 & \textbf{0.1} & 1.0 \\
        \midrule
        Acc. (\%)       & 78.6 & 78.6 & \textbf{78.6} & 71.4 
                        & 75.0 & \textbf{78.6} & 64.3 & 71.4 
                        & 71.4 & 71.4 & \textbf{78.6} & 71.4 \\
        Sol. Rate (\%)  & 71.4 & 61.9 & \textbf{71.4} & 50.0 
                        & 63.1 & \textbf{71.4} & 59.5 & 61.9 
                        & 64.3 & 69.0 & \textbf{71.4} & 61.9 \\
        Complexity      & 14.79 & 13.29 & \textbf{13.36} & 14.00 
                        & 14.57 & \textbf{13.36} & 14.00 & 14.21 
                        & 13.86 & 13.36 & \textbf{13.36} & 12.64 \\
        \bottomrule
    \end{tabular}
\end{table}

\section{Additional Ablation Analysis} \label{appendix:ablation_results}

\begin{table}[!h]
    \centering
    \small 
    \caption{Performance of \ours for each ablation on the Feynman dataset}
    \label{tab:ablation_symbolic_results}
    \begin{tabular}{lrr}
    \toprule
     & \multicolumn{2}{c}{\textbf{Symbolic Solution Rate (\%)}}\\
    \cmidrule(lr){2-3}
    \textbf{Algorithm} & $0.0\%$ & $10\%$ \\
    \midrule
    CADSR & 41.59  & \textbf{30.93} \\
    Drop DPE & 42.24 & 27.69 \\
    Drop BIC & \textbf{43.43} &  14.65 \\
    Drop DCT & 40.84 & 30.17 \\
    Drop GPRO & 37.18 & 27.16 \\
    SPL ($\eta = 0.8$) & 40.19 & 27.59 \\
    SPL ($\eta = 0.99$) & 41.06 & 15.73 \\
    SPL ($\eta = 0.9999$) & 40.09 & 0.32 \\
    TPSR ($\lambda = 0.1$) & 36.83 &  11.16 \\
    TPSR ($\lambda = 1$) & 34.62 & 10.71 \\
    TPSR ($\lambda = 10$) & 36.85 & 10.81  \\
    \bottomrule
    \end{tabular}
\end{table}

\subsection{Reward Functions}

SPL and TPSR introduced reward functions with regularization terms and tunable hyperparameters. SPL's reward function is specified in \eqref{eq:spl}, and TPSR's reward function is~\eqref{eq:tpsr}.
SPL introduced a scalar term $\eta^n$, where $n$ denotes the number of multiplication operators in the expression, and $\eta$ is a hyperparameter. $\eta$ controls the strength of the regularizer where $\eta=1$ means no regularization and $\eta < 1$ strengthens the regularazation. The authors of the SPL work found that tuning $\eta$ significantly impacts the performance, and they suggest a starting value of $\eta=0.99$.
TPSR introduced an additive term $\lambda \exp(\frac{-l(\tau)}{L})$, where $L$ is the max number of tokens, $l(\tau)$ is the length of the expression, and $\lambda$ is a hyperparameter. $\lambda$ controls the regularization, and having a large 
$\lambda$ will increase the regularization. 

\begin{align}
    R(\tau) = \frac{\eta^{n}}{1 + \sqrt{\frac{1}{N}\sum^{N}_{i=1}(y_i - \tau(\x_i))^{2}}} \label{eq:spl}\\
    R(\tau) = \frac{1}{1 + \text{NMSE}(y, \tau(\x))} + \lambda \exp(\frac{-l(\tau)}{L})\label{eq:tpsr}
\end{align}

\subsection{DCT Attention} \label{appendix:dct_runtimes}

The DCT converts the input into the signal space before using the attention mechanism, which maintains the transformer's time complexity of $\mathcal{O}(n^2 d)$, where $n$ is the sequence length and $d$ is the embedding dimension. We reduce the embedding space by clipping the highest frequency components. For SRBench, we clipped the 2 highest frequency signals from a 10-dimensional embedding space. Therefore, we should attain around a 20\% decrease in sampling time of the standard transformer. 
Since the runtime for the whole SRBench is dominated by the optimization of the constant token(s), to exclusively evaluate the sampling efficiency, we tested with the \texttt{Nguyen-4} problem, which does not include the constant token in the token library. We ran our method with DCT attention layers and with ordinary attention layers, each for 50 trials with 100 epochs per trial on an RTX 3080 GPU. We recorded the time each architecture took to predict tokens. The empirical results, as shown in Table \ref{tab:dct_runtimes}, match the theoretical runtime reduction, resulting in a $21\%$ decrease in sample time using the DCT attention layers.

\begin{table}[!h]
    \centering
    \small
    \caption{\small Sampling Time on \texttt{Nguyen-4} with 50 trials.}
    \label{tab:dct_runtimes}
    \begin{tabular}{ccc}
    \toprule
    Architecture & Total Sample Time (s) & Sample Time per Expression (ms) \\
    \midrule
    DCT attention & 1072.8 & 3.576 \\
    Standard attention & 1354.2 & 4.514 \\
    \bottomrule
    \end{tabular}
\end{table}

\subsection{GRPO}\label{appendix:grpo}

\begin{figure}
    \centering
    \includegraphics[width=\linewidth]{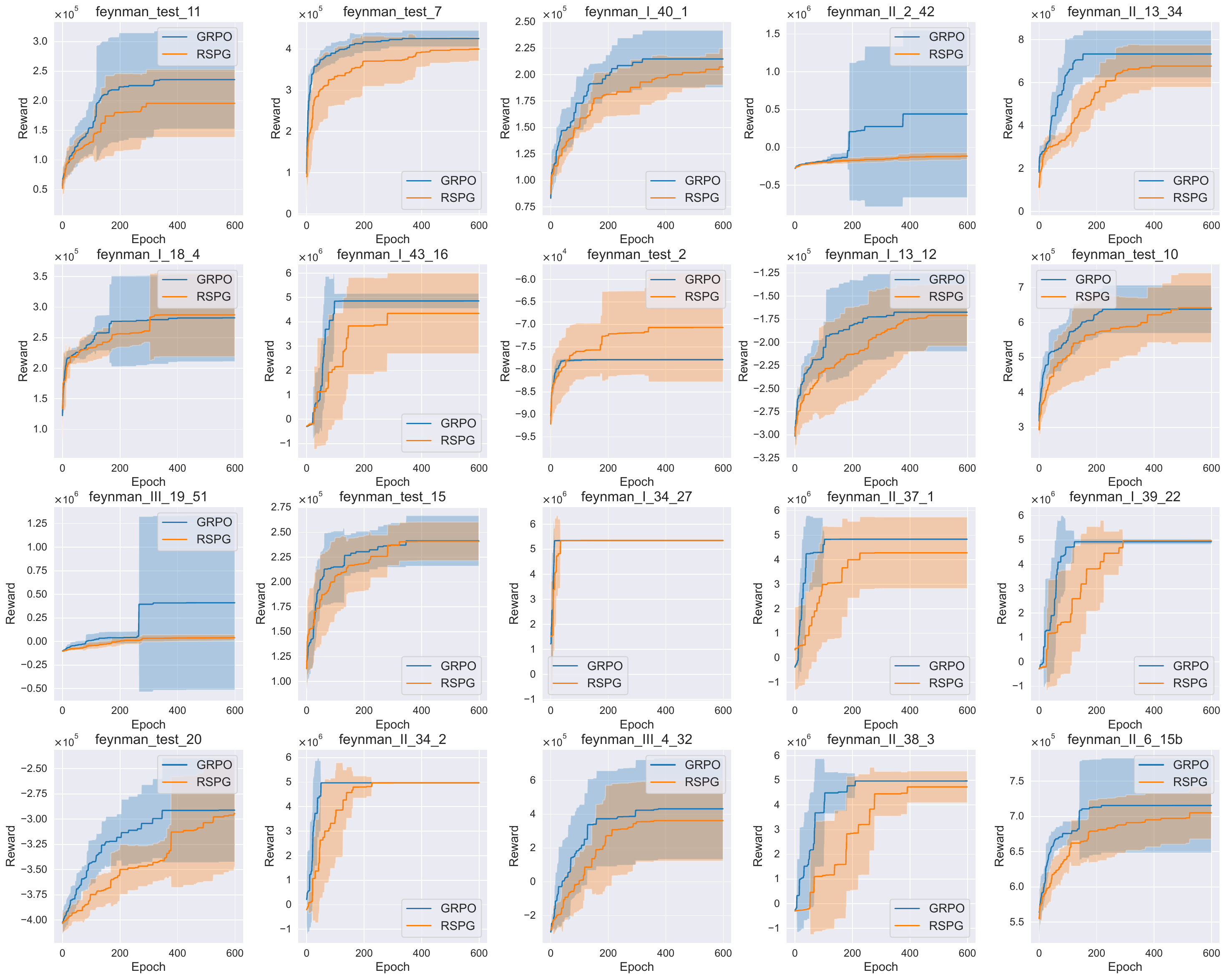}
    \caption{\small Twenty examples of learning curves (the number of epochs \textit{vs.} reward) for running \ours with GRPO and with standard risk seeking policy gradient (RSPG). The problems were  randomly selected from the Feynman dataset. Errors are one standard deviation.}
    \label{fig:Feynman_cumlatives}
\end{figure}

We observed that GRPO can increase performance and convergence rate compared to the standard risk-seeking gradient.  
Figure \ref{fig:Feynman_cumlatives} includes 20 random problems from Feynman dataset in SRBench. We can see in all the cases, GRPO enables the model to converge faster than the standard risk-seeking policy gradient. In the vast majority of the case, GRPO leads to a higher reward. 
On average, GRPO improves the symbolic recovery rate across the entire dataset. 

\subsection{Dual-Position Encoding Info}
In the DPE ablations, we observed that replacing DPE with a standard 1D position encoding results in a marginal decrease in symbolic recovery peformance at $0.0\%$ noise, while improving symbolic recovery at $10\%$ noise. We conject that this effect is realated to the improvement in exploration that the DPE introduces. Empirically, we found that using DPE consistently increases the diversity of expressions generated by the model compared with standard 1D positional encoding. We re-evaluated four randomly picked SRBench problems at the $10\%$ noise level with each positional encoding and found that DPE discovered an average of 473 novel expressions per epoch, whereas standard positional encoding discovered 284.

This increased exploration may function as a form of regularization: by reducing over-exploitation, it helps prevent the model from collapsing into poor modes — such as overly complex expressions that overfit noise — thereby improving robustness in noisy settings. Conversely, in the noise-free setting, this additional exploration may slightly hinder performance because the model may spend more effort exploring alternatives rather than converging quickly to the (easily discoverable) correct expression.

We suspect that DPE provides richer positional information — capturing both the depth and the horizontal position of tokens within the expression tree — which can lead to more expressive latent representations that enhance exploration. This, in turn, likely enables the policy to generate a broader set of candidate expressions. Evaluation of these statements poses challenges, as we can only observe correlations in the results, causing these statements to remain conjectures.

\subsection{Tail Barrier} \label{appendix:tail-barrier}
Empirically, we observed "tail barrier" effect occur fairly often as the reward distribution is often highly skewed, wherein the top $k$ expressions within the top-$\alpha\%$ receive a substantially larger reward than the rest. 
These top-$k$ expressions dominate the policy update due to the difference in reward, resulting in a "tail barrier" effect. Importantly, these top-$k$ expressions are not always representative or a derivative of the underlying expressions and can lead to a local optimum.

\begin{figure}[H]
    \centering
    \hspace{2.5cm}
    \begin{subfigure}[t]{0.33\linewidth}
        \centering
        \vspace{0pt}
        \includegraphics[width=\linewidth]{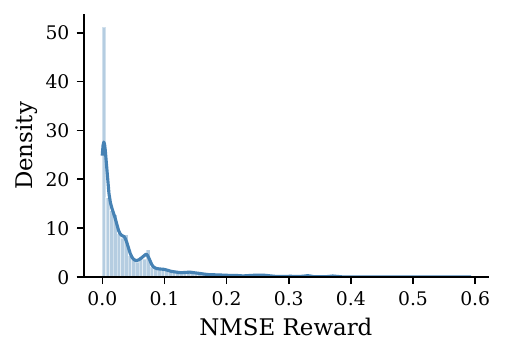}
        \caption{NMSE reward distribution during training}
        \label{fig:reward_distribution}
    \end{subfigure}
    \hfill
    \begin{subfigure}[t]{0.48\linewidth}
        \centering
        \vspace{3mm}
        \begin{tabular}{lc}
            \toprule
            Top-$k$ & Percent Dominated \\ 
            \midrule
            0      & 0.74\%    \\
            1      & 1.51\%    \\
            3      & 2.39\%    \\
            5      & 3.16\%    \\
            10     & 6.29 \%   \\
            \bottomrule
        \end{tabular}
        \vspace{7mm}
        \caption{Percent of epochs dominated by $k$ expressions}
        \label{tab:top_k_table}
    \end{subfigure}
    \caption{Overall caption for both panels}
    \label{fig:tail-barrir}
\end{figure}

To observe the phenomenon, we evaluated five randomly selected problems from the Feynman dataset with five trials per problem. For each problem, we ran \ours with the standard NMSE reward and recorded the reward for each expression in the top-$\alpha\%$ at every training step. Across all epochs, we observed that the bottom $10\%$ of all expression rewards was below $10^{-3}$, whereas the top $10\%$ of expressions were above 0.12 - a two orders of magnitude difference. Figure \ref{fig:reward_distribution} highlights how significantly the reward is right-skewed.

We observed that 0.7\% of all epochs had a full-tail barrier, resulting in all rewards for that epoch being zero. Additionally, we found that $k$ expressions dominated an epoch if the top-$k$ expressions contained over 80\% of the total reward for the epoch. Table \ref{tab:top_k_table} shows that there is a significant probability that multiple epochs are dominated by a small set of expressions, leading to the effects of a tail barrier being a probable occurrence during training.

\section{Theoretical Results} \label{appendix:theory}

\subsection{Proof of Lemma~\ref{lem:tail}}\label{sect:proof-tail}
    
\begin{proof}
Pick up any reward value $z_0$. Due to the continuity of the mapping $f$, for arbitrary $\epsilon>0$, there exists $\delta>0$ such that for all $z\neq z_0$, if $|z-z_0| < \delta$, then $|f(z) - f(z_0)|<\epsilon$. Let us take $\epsilon = \frac{1}{2}s$, where $s$ is the machine precision (\eg $2^{-32}$).  Since the reward function is continuous, we can find a set of distinct reward values $z_1, \ldots, z_M$ from $B(z_0, \delta(\epsilon))=\{z \in \text{dom } f, |z - z_0|<\delta(\epsilon)\}$. Let us look at the mapped rewards, $f(z_1), \ldots, f(z_M)$. For any $1 \le i, j \le M$, we have
\[
|f(z_i) - f(z_j)| = |f(z_i) - f(z_0) + f(z_0) - f(z_j)| \le |f(z_i) - f(z_0)| + |f(z_0) - f(z_j)| < \frac{s}{2} + \frac{s}{2} = s. 
\]
Therefore, there are no numerical difference among these mapped rewards, and they can create a tail barrier. 
\end{proof}

\subsection{Proof of Lemma~\ref{lem:robust}}~\label{sect:proof-robust}
\begin{proof}
For any set of BIC reward values, $\Scal = \{R(\tau^{(1)}), \ldots, R(\tau^{(B)})\}$, we denote the mapped reward values by $\widehat{\Scal} = \{\whatR_1, \ldots, \whatR_B\}$, where each $\whatR_j = f(R(\tau^{(j)}))$ ($1 \le j \le B$). We know that each 
\begin{align}
    \whatR_j = \lambda \cdot \text{ReLU}\left( 1- \frac{\{\tau^{(i)}: R(\tau^{(i)}) > R(\tau^{(j)}), 1 \le i \le B \}|}{ B\alpha/100}\right) 
\end{align}
where $R_\alpha$ is the $1 - \frac{\alpha}{100}$ quantile of the rewards in $\Scal$. Let $B_\alpha$ denotes the number of top $\alpha\%$ expressions. We have $B\alpha/100 = B_\alpha$, and 
\[
\whatR_j = \lambda \cdot \text{ReLU}\left( 1- \frac{\{\tau^{(i)}: R(\tau^{(i)}) > R(\tau^{(j)}), 1 \le i \le B \}|}{ B_\alpha}\right).
\]
Since for every top $\alpha\%$ expression $\tau^{(j)}$, the numerator $|\{\tau^{(i)}: R(\tau^{(i)}) > R(\tau^{(j)}), 1 \le i \le B \}|$ is an integer and is guaranteed to be less than $B_\alpha$, $\whatR_j$ is always positive.  
Accordingly, every expression in the top $\alpha$\% will not have their gradient zeroed out (see~\eqref{eq:new-risk-seeking}), and we will never meet a tail or partial tail barrier. 

To show the unbiasedness, we need to replace $R_\alpha$ by the $1-\alpha/100$ quantile of the distribution of the BIC reward $R$. Denote the mapped reward by $\whatR$. Since $\whatR$ is bounded, namely, $\whatR \in [0, 1]$, we can follow exactly the same steps in the proof of the original risk-adverse gradient paper~\citep{tamar_policy_2014} to show the unbiasedness of~\eqref{eq:new-risk-seeking}. 
\end{proof}

\section{SRBench}

\subsection{Symbolic Problems}

We provide Table \ref{tab:results_symbolic_regression} containing the numerical values for all of the algorithms for all of the algorithms we compare within Figure \ref{fig:ground_truth_sr_bench}. Bolded values are the best  for the given column. \ours has achieved the best performance for $0.1\%$, $1\%$, and $10\%$ noise, generating the second most simple models on average for these noise levels. MRGP has the best performance in terms of accuracy but generates overly complex expressions with high accuracy. Note that uDSR is omitted because its exact numerical data is not available. SRBench uses SymPy\footnote{\url{https://www.sympy.org/en/index.html}} to evaluate symbolic accuracy. However, SymPy is known for generating false negatives due to incomplete simplifications. For our table, we also counted any expression as being symbolically correct when the $R^2$ test score is exactly 1.0, as it is improbable that an expression that is not symbolically equivalent will perfectly predict at $\sim 20,000$ testing points. Additionally, we compare with DySymNet in Table \ref{tab:DySymNet_srbench} for SRBench at $0.0\%$ noise based on the ability categories reported in DySymNet's original paper.

\begin{table}[!h]
    \centering
    \scriptsize
    \caption{\small Symbolic regression performance in SRBench with known solutions.}
    \label{tab:results_symbolic_regression}
    \begin{tabular}{lrrrrrrrrrrrr}
    \toprule
     & \multicolumn{4}{c}{\textbf{Symbolic Solution Rate (\%)}} & \multicolumn{4}{c}{\textbf{Accuracy Rate (\%)}} & \multicolumn{4}{c}{\textbf{Simplified Complexity}} \\
    \cmidrule(lr){2-5} \cmidrule(lr){6-9} \cmidrule(lr){10-13}
    \textbf{Algorithm} & $0.0\%$ & $0.1\%$ & $1\%$ & $10\%$ & $0.0\%$ & $0.1\%$ & $1\%$ & $10\%$ & $0.0\%$ & $0.1\%$ & $1\%$ & $10\%$  \\
    \midrule
    AFP & 22.31 & 19.92 & 16.85 & 12.85 & 43 & 42 & 40 & 41 & 29.35 & 28.43 & 29.01 & 30.82 \\
    AFP\_FE & 28.08 & 22.69 & 20.31 & 12.85 & 56 & 50 & 50 & 50 & 34.58 & 35.66 & 34.45 & 36.77 \\
    AIFeynman & \textbf{61.84} & 31.89 & 12.61 & 0.86 & 74 & 74 & 68 & 10 & 83.29 & 88.66 & 99.27 & 110.54 \\
    BSR & 2.50 & 0.61 & 0.08 & 0.00 & 12 & 11 & 12 & 7 & 34.29 & 35.51 & 36.80 & 38.38 \\
    Bingo & 48.77 & 14.62 & 4.77 & 0.77 & 64 & 60 & 62 & 59 & 15.56 & 19.29 & 21.32 & 22.54 \\
    CADSR & 41.83 & \textbf{34.27} & \textbf{33.57} & \textbf{29.25} & 55 & 54 & 55 & 54 & 16.30 & 15.69 & 15.89 & 15.03 \\
    DSR-OCT & 24.81 & 24.42 & 17.53 & 10.48 & 38 & 41 & 40 & 39 & 16.57 & 16.10 & 16.96 & 18.45 \\
    DSR-NCT & 19.71 & 19.23 & 18.92 & 16.61 & 24 & 25 & 25 & 25 & \textbf{13.14} & \textbf{14.36} & \textbf{14.61} & \textbf{14.40} \\
    EPLEX & 12.50 & 9.92 & 8.77 & 9.54 & 44 & 45 & 52 & 47 & 53.24 & 51.74 & 49.91 & 40.04 \\
    FEAT & 0.10 & 0.00 & 0.00 & 0.00 & 40 & 43 & 41 & 14 & 88.01 & 77.32 & 72.61 & 50.40 \\
    FFX & 0.00 & 0.00 & 0.00 & 0.08 & 0 & 0 & 3 & 18 & 274.88 & 273.29 & 286.03 & 341.38 \\
    GP-GOMEA & 43.08 & 10.62 & 4.69 & 1.46 & 71 & 70 & 73 & \textbf{68} & 25.73 & 32.75 & 37.59 & 45.41 \\
    ITEA & 20.77 & 13.77 & 7.69 & 1.46 & 27 & 27 & 27 & 26 & 14.46 & 14.96 & 15.35 & 16.00 \\
    MRGP & 0.00 & 0.00 & 0.00 & 0.00 & \textbf{93} & \textbf{92} & \textbf{89} & 2 & 109.95 & 106.50 & 83.06 & 0.00 \\
    Operon & 16.00 & 12.31 & 1.92 & 0.08 & 87 & 86 & 86 & 73 & 40.80 & 40.13 & 60.40 & 70.78 \\
    SBP-GP & 22.69 & 0.69 & 0.00 & 0.00 & 74 & 74 & 75 & 54 & 109.94 & 102.09 & 112.93 & 116.30 \\
    TPSR & 36.09 & 0.00 & 0.00 & 0.00 & 68 & 68 & 66 & 38 & 57.11 & 59.32 & 63.42 & 65.83 \\
    gplearn & 16.15 & 16.86 & 16.59 & 16.00 & 30 & 29 & 27 & 22 & 45.80 & 37.76 & 36.42 & 33.84 \\
    \bottomrule
    \end{tabular}
\end{table}

\begin{table}
    \centering
    \caption{R$^2$ scores on SRBench with 0.0\% noise.}
    \label{tab:DySymNet_srbench}
    \begin{tabular}{lcccccc}
        \toprule
        Dataset & CADSR & DySymNet & GP-Gomea & Operon & TPSR & AIFeynman \\
        \midrule
        Feynman & 0.9363 & 0.9931 & \textbf{0.9969} & 0.9908 & 0.9925 & 0.9045 \\
        Strogatz & 0.9689 & 0.9968 & 0.9975 & \textbf{0.9994} & 0.9648 & 0.5900 \\
    \bottomrule
    \end{tabular}
\end{table}

\subsection{Black Box Problems} \label{appendix:blackbloxresults}

Table \ref{tab:blackbox_problems_table} shows the numerical performance metrics and runtimes of each method tested on the black box SRBench dataset. While \ours does not give the best metric in any category, it does appear on the Pareto front in Figure \ref{fig:blackbox-srbench}. Notably, SBP-GP, TPSR, and Operon are the best-performing algorithms in terms of $R^2$ test score on the black box problems and have a higher level of complexity than \ours. SBP-GP and Operon are GP-based methods. TPSR is a transformer that uses MCTS on a pretrained transfomer to predict lengthy and accurate expressions.

\begin{table}[!h]
    \centering
    \small
    \caption{\small Symbolic regression performance and runtime on black box problems in SRBench.}
    \label{tab:blackbox_problems_table}
    \begin{tabular}{cccc}
    \hline
    \textbf{Algorithm} & \textbf{R² Test} & \textbf{Model Size} & \textbf{Training Time (s)} \\
    \hline
    AFP & 0.657613 & 34.5 & 6002.55 \\
    AFP\_FE & 0.664599 & 35.6 & 6153.50 \\
    AIFeynman & -3.745132 & 2500 & 82069.31 \\
    AdaBoost & 0.704752 & 10000 & 65.29 \\
    BSR & 0.257598 & 19.8 & 20426.86 \\
    Bingo & 0.711951 & 22.2 & 53537.42 \\
    CADSR & 0.615459 & 11.9 & 5643.43 \\
    DSR-NCT & 0.571669 & \textbf{8.89} & 35096.16 \\
    DSR-OCT & 0.642417 & 14.8 & 4267.68 \\
    EPLEX & 0.760414 & 55.8 & 15673.69 \\
    FEAT & 0.784662 & 74.2 & 6723.91 \\
    FFX & -0.667716 & 1570 & 243.74 \\
    GP-GOMEA & 0.746634 & 27.3 & 9063.54 \\
    ITEA & 0.640731 & 112 & 12337.48 \\
    KernelRidge & 0.615147 & 1820 & 38.88 \\
    LGBM & 0.637670 & 5500 & 28.53 \\
    Linear & 0.454174 & 17.4 & \textbf{0.24} \\
    MLP & 0.531249 & 3880 & 30.47 \\
    MRGP & 0.417864 & 12100 & 190697.78 \\
    Operon & 0.794831 & 65.0 & 2979.43 \\
    RandomForest & 0.698541 & 1.54e+06 & 120.21 \\
    SBP-GP & \textbf{0.798932} & 639 & 166745.46 \\
    TPSR & 0.792001 & 95.7 & 504.38 \\
    XGB & 0.775793 & 16400 & 240.58 \\
    gplearn & 0.541264 & 16.3 & 24157.99 \\
    \hline
    \end{tabular}
\end{table}

\section{Run Times}\label{appendix:runtime}
\begin{figure}[!h]
    \centering
    \includegraphics[width=0.75\linewidth]{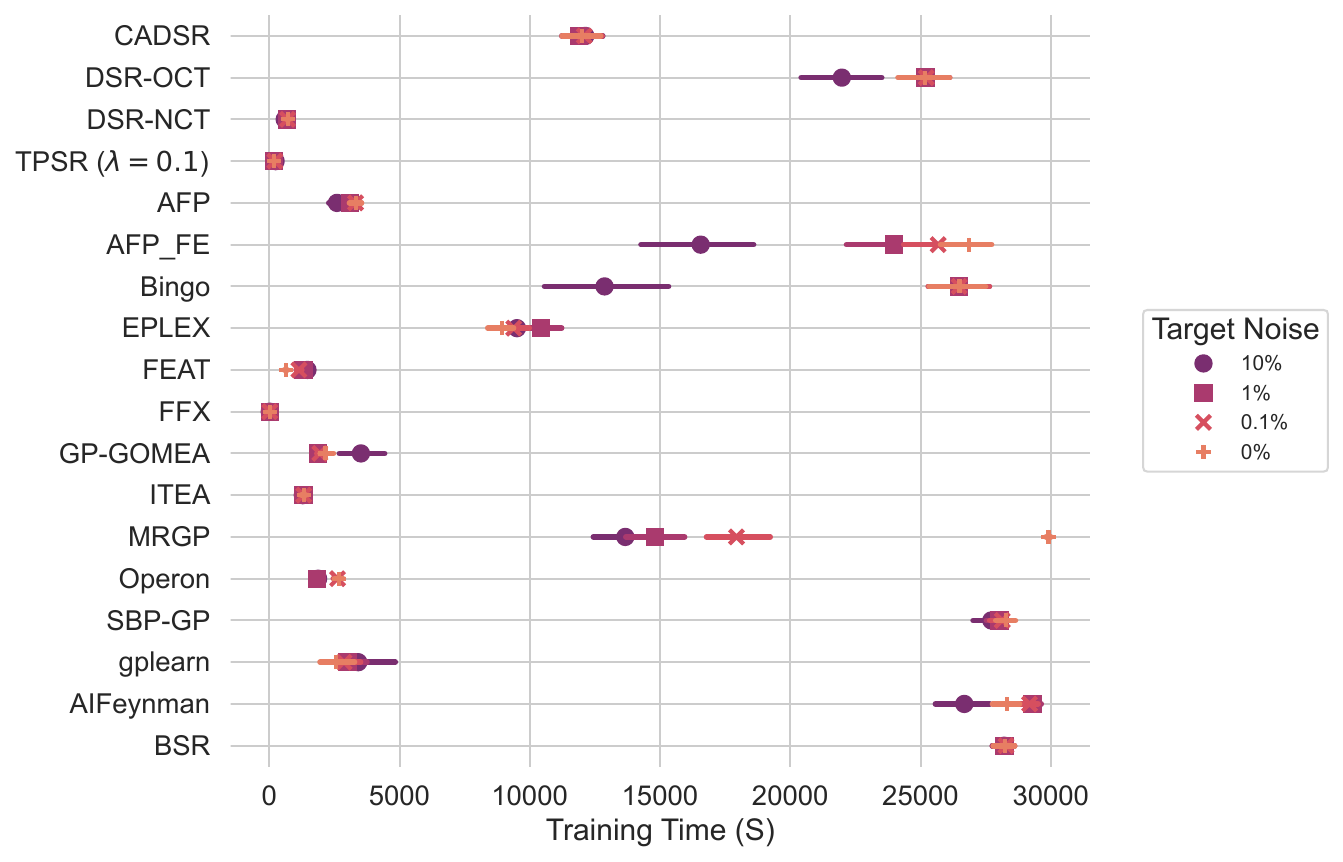}
    \caption{\small Runtime in solving SRBench problems with solutions known. Error bars denote a $95\%$ confidence interval.}
    \label{fig:srbench_time_comparison}
\end{figure}

Figure \ref{fig:srbench_time_comparison} shows the recorded runtime of each method in solving  the symbolic problems from SRBench. However, this is a rough comparison, since \ours does not use any early stopping criterion and the hardware resources --- though comparable --- are not exactly the same across all the methods for this benchmark. 

\subsection{Expressions Evaluated}
Evaluated expressions are an additional metric to compare symbolic regression methods. Table \ref{tab:evaluated_expressions} compares the average number of expressions evaluated by \ours and DSR across 5 trials of the Stragatz dataset. In these trials \ours had a 4-hour runtime limit, and DSR had an 8-hour runtime limit, and \ours had an evaluation budget of 600k while DSR had an evaluation budget of 500k. \ours achieved 40\% reduction in evaluations compared to DSR on noisy datasets while maintaining higher accuracy and a higher symbolic recovery rate. These results suggest that \ours is more sample-efficient in identifying notable expressions. In the majority of the experiments, DSR exhausts its evaluation budget, while \ours terminates earlier due to runtime limitations. Notably, \ours had a higher evaluation budget of 600k, but rarely surpassed 450k evaluations during these trials.

\begin{table}[h]
    \centering
    \caption{Evaluated expressions for DSR and \ours on the Strogatz dataset}
    \begin{tabular}{cccc}
    \hline
    Noise Level & DSR & CADSR & Reduction (\%) \\
    \hline
    0.0\% & 338k & 331k & 2.1 \\
    0.1\% & 500k & 287k & 42.6 \\
    1.0\% & 500k & 272k & 45.6 \\
    10.0\% & 500k & 299k & 40.2 \\
    \hline
    \end{tabular}
    \label{tab:evaluated_expressions}
\end{table}

\section{Fracture Mechanics}\label{sect:fracture}

The dataset contains the following 6 features: Max Schmid of Grain, Average Schmid Factor of Grain, Number of High Grain Schmids, Principal Components Grain, Schmid Variance in Grain, sin NST. Complexity of each expression was calcualate by counting the number of nodes in the expression tree using SymPy.

Table \ref{tab:crack_full_table} comparing average accuracy, runtime, and complexity of the four methods. Each method was given access to up 10 CPU cores and a GPU. TPSR was ran with the default parameter settings for the model on a H200 GPU, as the model required over 20 GB of VRAM. DSR and GP-GOMEA used the same parameters used for the SRBench benchmark. DySymNet was given an 8 hour runtime with the same GPU and CPU resources as \ours. Treating this as a real-world problem, we ran \ours without the hardware limitations used for SRBench, enabling potential improvements in runtime and performance. A list of hyperparameters for \ours can be found in Table \ref{tab:hyperparameters}. Table \ref{tab:best_expressions} reports the best overall expression found by each method. \ours achieves similar accuracy as GP-GOMEA, while providing the most human interpretable expression. DySymNet is excluded from Table \ref{tab:best_expressions} due to its high complexity making it unsuitable for print.

To ensure a fair comparison, we tested DSR with a batch size of 2000, as to match the batch size of \ours. We labeled this experiment DSR (2000) in Table \ref{tab:crack_full_table}. The larger batch size caused all eight DSR trials to converge early, leading to identical expression discovered in across all trials.
While this version of DSR discovers an expression that is less complex than \ours's expression, it significantly impacts $R^2$ score.

\begin{table}[h!]
    \centering
    \small
    \caption{\small Symbolic regression performance for predicting crack initiation.}
    \begin{tabular}{lrrrr}
        \hline
        Method   & Train $R^2$ & Test $R^2$ & Runtime (s) & Complexity \\
        \hline
        CADSR      &  0.636  &  0.626  &  11583.643 &  22.250  \\
        DSR        &  0.622  &  0.616  &  7479.159  &  24.625  \\
        DSR (2000) &  0.590  &  0.587  &  1871.127  &  21.000  \\
        TPSR       &  0.043  &  0.046  &  667.433   &  40.750  \\
        GP-GOMEA   &  0.645  &  0.636  &  7479.687  &  73.000  \\
        DySymNet   &  0.584  &  0.456  &  23766.691 &  1696.875 \\
        \hline
    \end{tabular}
    \label{tab:crack_full_table}
\end{table}

\begin{table}[h!]
    \centering
    \small 
    \renewcommand{\arraystretch}{1.5} 
    \caption{\small Top expressions found by each method for predicting crack initiation.}\label{tab:expr-fracture}
    \begin{tabular}{lccc}
        \hline
        Method & $R^2$ & Complexity & Simplified Expression \\
        \hline \hline
        CADSR & 0.663 & 21  &  
        $c_0 + c_1 x_0 + \tfrac{c_2}{x_3} + \tfrac{c_3 x_4 + c_4}{x_1}$ \\[0.4em]
        \hline
        
        DSR   & 0.634 & 26 &  
        $-\tfrac{x_2 \Big(\exp(x_5 \log(\cos(x_1)) \cos(x_2)) - c_0 \sin(c_1 x_1 - x_5 + c_2)\Big)}{x_4}$ \\[0.4em]
        \hline

        DSR (2000) & 0.616 & 21  & 
        $-\tfrac{\sin\!\Big(x_1 (x_1 + x_4) \exp(c_0 x_1 + c_1 x_2) + c_2\Big)}{x_4}$ \\[0.4em]
        \hline

        TPSR & 0.372 & 45  & 
        \begin{tabular}[c]{@{}l@{}}
        $c_{0} - c_{1} \Big( \Big(c_{2} - c_{3} x_0\Big) \cdot \Big(c_{4} \, |c_{5} x_1 - c_{6} x_3 + c_{7}| - c_{8}$ \\[0.3em]
        $\quad - \tfrac{c_{9}}{c_{10}\sin(-c_{11} x_2 + c_{12} x_4 + c_{13} x_5 + c_{14}) - c_{15}} \Big)\Big)^{0.5}$
        \end{tabular} \\[0.8em]
        \hline

        GP-GOMEA & 0.668 & 87 & 
        \begin{tabular}[c]{@{}l@{}}
        $c_{0} + c_{1} \Big(-\sin(c_{2} x_2) + c_{3} \tfrac{(x_1 + x_4)}{x_3}\Big)$ \\[0.3em]
        $\quad \cdot \Big(c_{4} x_2 + c_{5} x_3 - x_4 + c_{6} x_5 - \cos(x_0 + x_1 - x_3 x_5)\Big)$ \\[0.3em]
        $\quad \cdot \tfrac{\cos(x_0 - x_4)}{\sqrt{-x_0 + x_4}} + \tfrac{c_{7} \cos(x_0/x_2) \cos\!\Big(\tfrac{-x_1 - c_{8}}{\sqrt{x_1}}\Big)}{x_4}+ \tfrac{c_{9} \cos(c_{10}/\sqrt{x_3})}{x_3}$
        \end{tabular} \\[0.8em]
        \hline
    \end{tabular}
    \label{tab:best_expressions}
\end{table}